\documentclass[12pt]{article}
\usepackage{amsmath}
\usepackage{graphicx}
\usepackage{natbib}
\usepackage{url} 
\usepackage[shortlabels]{enumitem}
\usepackage{latexsym, epsfig, amssymb, amsthm, mathrsfs, bbm, enumitem}
\usepackage[OT1]{fontenc}
\usepackage{xcolor}
\usepackage[colorlinks,citecolor=blue,linkcolor=blue,urlcolor=blue]{hyperref}
\usepackage{multirow,multicol,booktabs}
\usepackage{anysize}
\usepackage[left=1in, right=1in, top=1in, bottom=1in]{geometry}%

\usepackage[utf8]{inputenc} 
\usepackage{url}            
\usepackage{amsfonts}       
\usepackage{nicefrac}       
\usepackage{microtype}      
\usepackage{comment}
\usepackage{caption}
\usepackage{enumitem}
\usepackage{etaremune}

\makeatletter

\usepackage{tikz}
\usepackage{float}
\usepackage{booktabs}
\usepackage{adjustbox}
\usepackage{algorithm}
\usepackage{algpseudocode}

\renewcommand{\theequation}{
	\arabic{equation}%
}

\newcommand{\ignore}[1]{}{}

\def\independenT#1#2{\mathrel{\setbox0\hbox{$#1#2$}%
		\copy0\kern-\wd0\mkern4mu\box0}}

\renewcommand{\ldots}{\cdots}

\renewcommand{\hat}{\widehat}

\usepackage{xcolor}
\usepackage[draft,inline,nomargin,index]{fixme}
\fxsetup{theme=color,mode=multiuser}
\FXRegisterAuthor{yf}{ayf}{\color{magenta}}

\usepackage{ulem}

\begin{document}

\title{LIDS: LLM Summary Inference Under the Layered Lens%
\thanks{
Dylan Park is Ph.D. Candidate, Department of Mathematics, University of Southern California, Los Angeles, CA 90089 (E-mail: \textit{dylanpar@usc.edu}). %
Yingying Fan is Centennial Chair in Business Administration and Professor, Data Sciences and Operations Department, Marshall School of Business, University of Southern California, Los Angeles, CA 90089 (E-mail: \textit{fanyingy@marshall.usc.edu}). %
Jinchi Lv is Kenneth King Stonier Chair in Business Administration and Professor, Data Sciences and Operations Department, Marshall School of Business, University of Southern California, Los Angeles, CA 90089 (E-mail: \textit{jinchilv@marshall.usc.edu}). %
}
\date{February 17, 2026}
\author{Dylan Park, Yingying Fan and Jinchi Lv
	\medskip\\
	University of Southern California
	\\
} 
}

\maketitle

\begin{abstract} 
Large language models (LLMs) have gained significant attention by many researchers and practitioners in natural language processing (NLP) since the introduction of ChatGPT in 2022. One notable feature of ChatGPT is its ability to generate summaries based on prompts. Yet evaluating the quality of these summaries remains challenging due to the complexity of language. To this end, in this paper we suggest a new method of LLM summary inference with BERT-SVD-based direction metric and SOFARI (LIDS) that assesses the summary accuracy equipped with interpretable key words for layered themes. The LIDS uses a latent SVD-based direction metric to measure the similarity between the summaries and original text, leveraging the BERT embeddings and repeated prompts to quantify the statistical uncertainty. As a result, LIDS gives a natural embedding of each summary for large text reduction. We further exploit SOFARI to uncover important key words associated with each latent theme in the summary with controlled false discovery rate (FDR). Comprehensive empirical studies demonstrate the practical utility and robustness of LIDS through human verification and comparisons to other similarity metrics, including a comparison of different LLMs.
\end{abstract}
	
\textit{Running title}: LIDS
	
\textit{Key words}: Large language models; Natural language processing; Summary inference; BERT; Latent SVD; Large text reduction; SOFARI and FDR control

\section{Introduction} \label{new.Sec.intro}

The advent of the Transformer \citep{transformer2017} has revolutionized the field of natural language processing (NLP) and led to the developments of the large language models (LLMs) such as ChatGPT \citep{ChatGPT2022,Roumeliotisetal2023} first released in 2022 as well as Claude \citep{Claude2023}, 
DeepSeek \citep{DeepSeek2025}, 
Gemini \citep{Gemini2023}, 
Grok \citep{Grok2024}, 
and Llama \citep{Touvronetal2023}. 

Thanks to the leap in the massive scale of the LLM in terms of the number of model parameters trained over an enormous collection of texts (including the internet generated contents), LLMs such as ChatGPT exhibit unprecedented power of large text comprehension and token generation, much beyond the simple capability of memorization. 

There is a rapidly growing literature on the Transformer-based LLMs, with various extensions of the original Transformer architecture including the BERT model for enhancing the efficiency, scalability, and general performance \citep{Devlinetal2018,Raffel2019,Fedus2021}. There are works on optimizing the Transformer model size determining scaling laws \citep{Kaplan2020}, optimal training \citep{Hoffmann2022}, and balancing power and computational efficiency \citep{Dao2022}. It has been explored how Transformers develop surprising capabilities including in-context learning without fine-tuning \citep{Brown2020} and observing new capabilities that emerge unpredictably as model size increases \citep{Wei2022}. 
Transformer has also been extended to multimodal inputs with a vision-language Transformer trained on image-text pairs \citep{Radford2021} as well as advanced image generation using Transformers conditional on text \citep{ramesh2022}. Retrieval-augmented generation (RAG) allows Transformers to retrieve and incorporate external knowledge during generation to help prevent hallucinations in LLMs \citep{Lewis2020}. Applications of LLMs in such areas as chemistry and finance were discussed in \citet{Ramosetal2024,Lietal2023,Nieetal2024,Yangetal2024,Kaddouretal2023}. 
For more comprehensive reviews of some recent developments on LLMs, see, e.g., \citet{Minaeeetal2024,
Naveedetal2023,Guoetal2023}. 

Statistical learning and inference for the LLMs have also received growing amount of attention in recent years. To name a few, \cite{Cherianetal2024} proposed new conformal inference methods that can provide conditionally valid filtering of claims from the LLM outputs, ensuring a high-probability guarantee of correctness in the identified subset of text. \cite{Lietal2025} introduced a general statistical framework based on hypothesis testing for LLM-generated text detection (i.e., watermarking), with theoretical guarantees. \cite{Chanetal2025} proposed the conformal information pursuit for evaluating the sequential information gain based on conformal prediction sets that can interactively guide the LLMs with shorter query-answer chains. \cite{Davidovetal2025} introduced a framework for quantifying the time-to-unsafe-sampling based on survival analysis and conformal prediction that provides provably calibrated predictive lower bounds on the time-to-unsafe-sampling of a given prompt. \cite{FanLvSunWang2025} suggested the LLM-TS, an LLM-based method for time series prediction inference incorporating online text data, showcasing that the LLM-generated surrogates of inflation could be exploited to construct tighter prediction intervals for inflation forecasting in economics.

Despite the existing works on the LLMs, statistical inference for the LLM summary remains largely unexplored. The user-specified prompts provide an effective channel of communications and interactions with the LLM, with the outputs typically generated in a random fashion from repeated open-ended prompts or some 
tweaks. In particular, 
summary is a powerful functionality of the LLM, serving as a fundamental tool for large text reduction. Such large non-numeric (i.e., text) data reduction is similar in flavor to large data reduction for numeric data using, e.g., the principal component analysis (PCA) or singular value decomposition (SVD), in the sense that the key themes in the original text should ideally be retained in the condensed summary.

Given the complicated nature of language, it remains largely unclear how to evaluate the accuracy and uncertainty of the LLM summary of some given large text in a statistically principled fashion. First, it is natural to expect that different LLMs will produce different versions of the summary, including the use of words and the structure of the writing. Second, the use of the same prompt for an LLM such as ChatGPT may result in a (slightly) different realization of the random summary when the prompt is repeated. One simple example is that a prompt specifying the desired length of the summary may not yield a summary with the exact length. Third, it is also to the benefit of the LLM when a randomized version of the LLM summary is presented so that the user feedback through follow-up prompts could provide meaningful data for refining the pre-trained LLM at a later stage.

To address the aforementioned practical challenges, we suggest in this paper a new framework of LLM summary inference with BERT-SVD-based direction metric and SOFARI (LIDS). Such framework quantifies the accuracy and statistical uncertainty associated with the LLM-generated summary, enabling us to gain insights into different LLMs on the summary functionality, the summary text embeddings, and the important key words under the layered lens. For a given text, we first apply the BERT model \citep{Devlinetal2018} to construct the  token embedding vectors that take into account the token meanings, contexts, and orderings. We then perform the SVD (or the sparse SVD if needed) to the BERT  embedding matrix and obtain an estimate of the latent SVD layers. We will introduce a novel LIDS direction metric based on the estimated SVD layers to quantify the closeness between the summary and the original text, through maximizing over the number of layers. The statistical uncertainty is characterized via repeated prompts. A natural byproduct from the first step of LIDS is the overall LIDS embedding vector obtained for the summary that is associated with the LIDS similarity measure, enabling large text reduction. One crucial question is how to unveil individual sets of key words corresponding to different latent themes given by the latent SVD layers with controlled error rate. To this end, we couple in the second step of LIDS the SOFARI \citep{zheng2023} for the latent SVD inference and the Benjamini--Hochberg (BH) procedure \citep{benjamini1995controlling} for the false discovery rate (FDR) control to identify sets of important key words for the layered themes.

\subsection{Related works} \label{new.sec.relawork}

There exist different ways in which the similarity between texts can be measured. Examples include the recall-oriented understudy for gisting evaluation (ROUGE) \citep{Lin2004}, metric for evaluation of translation with explicit ordering (METEOR) \citep{BanerjeeLavie2005}, bilingual evaluation understudy (BLEU) \citep{Papinenietal2001}, 
and Doc2Vec \citep{MikolovChenetal2013,LeMikolov2014, MikolovSutskever2013}. 
Those similarity measures often involve counting the frequencies of the same words and similar phrases. In practice, when comparing the voice of two texts, examining the frequency of similar phrases and words can be effective. Yet, such methods can fail when comparing the similarity of the content of texts. For instance, the two sentences ``The very rich man resides in a lavish and large home'' and ``The man lives in a mansion'' are very similar in meaning. However, if we determine a similarity score based on word frequencies, we will obtain a relatively low score, because the words used are mostly different. Thus, words and phrases with similar meaning should be rewarded in evaluating the similarity between two texts. Indeed, it is often the case that in order to summarize effectively and shorten the text without losing information, it is necessary to change the words used to get the same meaning. As seen in our two sentences example, the second sentence is almost half the length of the first one and uses different words with essentially the same information. Similarly for a large text, a high-quality summary may not include the identical words.

Another weakness in comparing texts by using words frequencies is that one may get false positives in similarity where texts that are wildly different in meaning, but use most of the same words and phrases. 
Examples of such texts could be generated in a MadLibs manner, where most words except the nouns, verbs, and adjectives are kept as constant between texts. For example, using the template ``The [adjective] [noun] decided to [verb] at the [place],'' we can generate two sentences that are completely different despite using many of the same words in the same order. Sentence 1: ``The happy dog decided to run at the park.'' Sentence 2: ``The angry teacher decided to yell at the classroom.'' These sentences would receive an inflated similarity score under the frequency measure despite being about completely different things. 
One common solution is to ignore common words that do not contribute to the meaning of the text, 
as many of such words are needed for conventional conversation and grammatical structure but alone contribute little to no information about the content matter.

To alleviate the aforementioned issues, one notable approach is the BERTScore \citep{Zhangetal2019BERTScore} based on the BERT embeddings of words for text generation evaluation. The use of the BERT embeddings of words naturally incorporates the rich information on the meanings, contexts, and orderings of words in the text. Similar to BERTScore, our LIDS similarity measure also builds upon the BERT embeddings of words, which is more flexible and robust than counting the frequencies of words and phrases. However, the LIDS similarity measure is fundamentally distinct from BERTScore in that BERTScore involves a (weighted) average of token-specific measure in terms of the maximum cosine similarity between token pairs, while LIDS involves the cosine similarity between a pair of singular values and singular vectors weighted tokens that incorporate the themed semantic meanings through the latent SVD layers.

\subsection{Our innovations} \label{new.sec.ourinno}

Our LIDS direction metric is an innovative text similarity measure for the LLM summary inference compared to the popularly used BERTScore due to the completely different ordering of the cosine similarity and the weighting, as mentioned in Section \ref{new.sec.relawork}. In particular, for the LIDS similarity measure, the weighting of tokens is done in a layered fashion through the singular values and singular vectors before the cosine similarity is calculated. In contrast, BERTScore applies the weighting after the maximum cosine similarities between token pairs are calculated.

The weights of inverse document frequency scores (idf) used in BERTScore are nonnegative, whereas the weights used in LIDS for each latent SVD layer are more flexible (i.e., both nonnegative and negative) due to the components of the left singular vectors and the signs of the right singular vectors. Further, the latent sparse SVD structure provides possibly different sets of key words across different themes corresponding to different SVD layers that are weighted by the corresponding singular values reflecting the contribution of that specific layer to the overall text (i.e., original or summary text). The FDR control on the sets of selected key words based on the p-values of the left singular vector components from SOFARI enables us with a layered lens revealing the important key words in the latent themes of the text with inference guarantees, which is largely lacking in the existing approaches.

Naturally, LIDS provides \textit{overall} text embeddings for large text reduction in contrast to the token-level embeddings used in BERTScore. In this sense, LIDS is more flexible than BERTScore and provides layered (i.e., themed) view of summary evaluation, where the subtle semantic meanings of text are captured through the singular values (i.e., the importance of each underlying theme) and the singular vectors (i.e., the important key words of each theme). Moreover, the nonnegative weights used in BERTScore reflect the typical word-piece token frequencies (i.e., the counting nature), whereas LIDS allows for more flexible weighting using both the signs and scalars (i.e., the singular values and singular vector components) in that we could now \textit{formally} add and subtract words, and multiply words by factors.
Through comprehensive text-based empirical studies, we systematically evaluate the efficacy of the LIDS similarity measure by investigating two benchmark summary mechanisms, human verification, comparisons to other similarity metrics, visualization for summary inference, comparison of different LLMs, and robustness checks over varying text domains. 

\begin{figure}[t]
    \centering
    \includegraphics[width=0.95\textwidth]{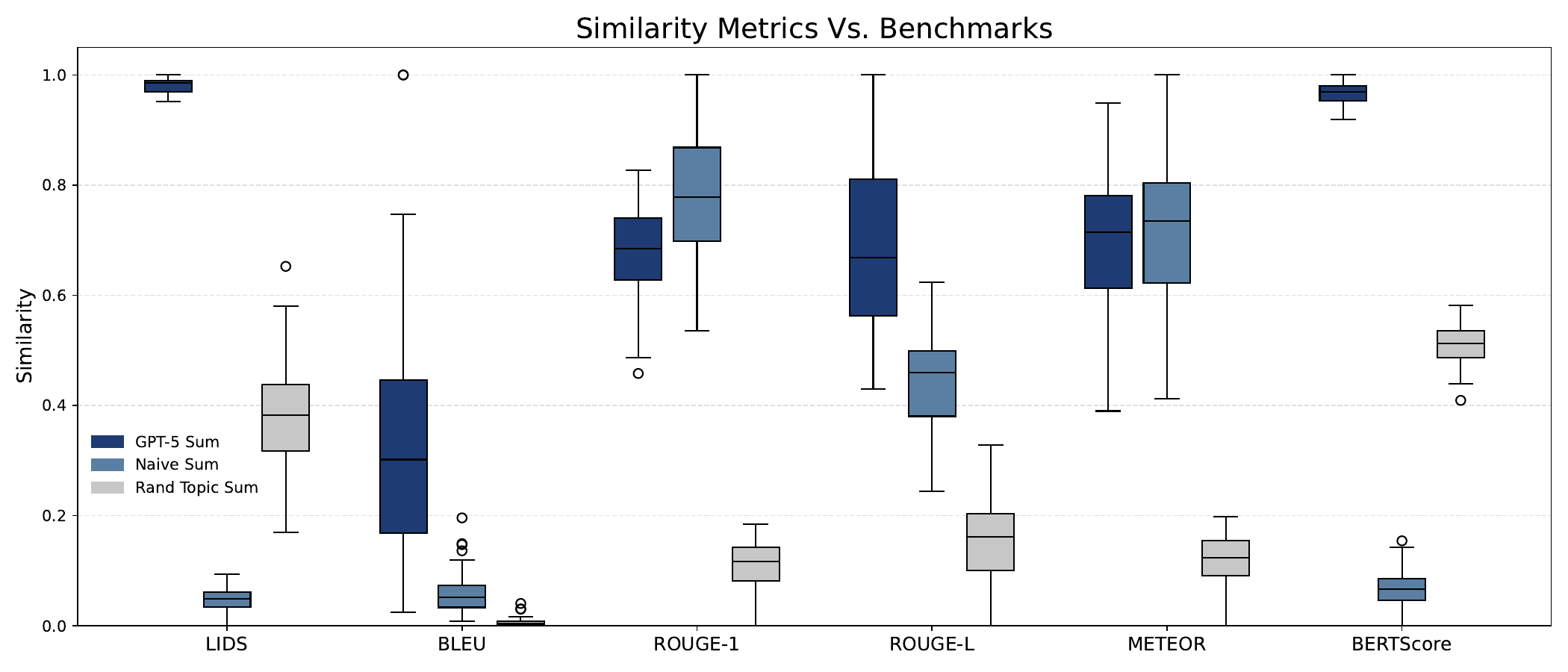}
    \caption{Rescaled boxplots of the LIDS, BLEU, ROUGE-1, ROUGE-L, METEOR, and BERTScore similarity measures for GPT-5 (dark blue) over $50$ repeated prompts and two benchmark summary mechanisms (light blue and gray) over $50$ random repetitions on the Utah article.
    }
    \label{fig:SimMets_Vs_Bench_Utah}
\end{figure}

As a preview of the empirical results presented later in Section \ref{new.Sec.empistud}, we compare the performance of the LIDS similarity metric against several established benchmark metrics such as the BLEU, ROUGE-1, ROUGE-L, METEOR, and BERTScore. Because these metrics operate on different scales, a direct comparison between their values is not sensible. However, it is still useful to examine how effectively each metric distinguishes the GPT-5 summaries from the two benchmark sets. To hold a side-by-side comparison, we rescale the outputs of each metric for the three summary sets to the interval $[0,1]$ using a linear transformation, mapping the minimum observed similarity to $0$ and the maximum one to $1$. Figure \ref{fig:SimMets_Vs_Bench_Utah} shows that LIDS and BERTScore outperform the other metrics by clearly evaluating the GPT-5 summaries with scores higher than either benchmark summary mechanism, with no overlap between the GPT-5 summary scores and any of the benchmark scores, clearly demonstrating an ability to differentiate between high-quality summaries and lower-quality ones. We also observe that LIDS performs 
better than BERTScore with much lower computation costs (See Tables \ref{table:Comp_Time_50_Sums} and \ref{table:Mem_Use}). Some additional benefits of LIDS over BERTScore will be discussed later when the context is clear.

The rest of the paper is organized as follows. Section \ref{new.Sec.directionrepprompts} introduces the latent BERT-SVD-based LIDS direction metric as well as the induced overall LIDS summary text embeddings for the first step of the LIDS algorithm, and two benchmark summary mechanisms. We present the second step of the LIDS algorithm on the FDR control for layered key word selection with SOFARI in Section \ref{new.Sec.sofarivisual}. Section \ref{new.Sec.empistud} provides a series of text-based empirical studies to illustrate the practical performance of LIDS for the LLM summary inference. We discuss some implications and extensions of our work in Section \ref{new.Sec.discuss}. The Supplementary Material provides the additional details of the implementation for LIDS and additional empirical results.

\section{A latent BERT-SVD-based LIDS direction metric and LIDS summary embeddings} \label{new.Sec.directionrepprompts}

In this section, we will introduce a latent BERT-SVD-based direction metric for the first part of the LIDS framework, the associated LIDS summary text embeddings, and two benchmark summary mechanisms.
\subsection{A latent BERT-SVD-based direction metric for LIDS} \label{new.sec.dirmet}

Assume that we are given an original text $T_0$, which we name as the reference text throughout the paper. The reference text can be any relatively large text (e.g., over $1000$ words or tokens), where the LLM summary for large text reduction is desired. 
The reference text can be a news article, a legal document, a book, or some other text. To illustrate the main ideas of the LIDS framework, we will focus on a Utah article presented in Section \ref{new.Sec.empistud} and further conduct robustness checks of LIDS over different text domains in Section \ref{Subsec.DocDomains}. Further assume that we are given a set of $m$ randomly generated LLM summaries of the reference text $T_0$ based on repeated prompts specifying the desired length of the summary. To ease the presentation, we will term each LLM summary as a test text hereafter (which is significantly shorter than the reference text) and denote by $\{T_j: 1 \leq j \leq m\}$ the resulting set of $m$ randomly generated LLM summaries of the reference text $T_0$. A main objective of the paper is to evaluate the similarity between each pair of reference text and test text $(T_0, T_j)$ with $1 \leq j \leq m$ and characterize the statistical uncertainty of such similarity measure via repeated prompts.

In implementation, we remove past prompts for each new LLM summary to avoid unintended biases in consecutive summaries. We may add a constraint that specifies the desired maximum and minimum word counts. In particular, we employ GPT-5 from ChatGPT as our motivating LLM to randomly generate $m = 50$ summaries using the prompt ``\texttt{Summarize the following text.}'' right before the reference text. We allow GPT-5 to determine the appropriate length of the LLM summary, giving rise to LLM summaries that generally range from $100$ to $200$ words in length. For each reference or test text, we exploit the BERT model \citep{Devlinetal2018} for construction of flexible token embedding vectors reflecting the underlying meanings, contexts, and orderings of tokens. Specifically, we adopt the Python package \texttt{BERT} to create a $p$-dimensional embedding vector for each token in the reference or test text. For our empirical experiments in Section \ref{new.Sec.empistud}, the BERT embedding dimensionality is set as the typical choice of $p = 768$. 

With the constructed BERT token embeddings, for each reference or test text $T_j$ with $j\in \{0\} \cup \{1,\ldots, m\}$, we can obtain a BERT embedding matrix $X_j \in \mathbb R^{n_j\times p}$ for text $T_j$, where $n_j$ denotes the number of tokens in text $T_j$ and each row represents a $p$-dimensional BERT embedding vector of the corresponding token in the text. In implementation, one may impose additional token restrictions such as zeroing out rows in $X_j$ that correspond to stop words to reduce the noise and influence of less important words in the text if desired. To provide a layered view of each reference or test text $T_j$ with $0 \leq j \leq m$, we construct estimates of the latent SVD layers using the corresponding BERT embedding matrix $X_j \in \mathbb R^{n_j\times p}$. For computational efficiency, we adopt the Python package \texttt{Numpy} to calculate the SVD directly. When the sparse SVD structure is desired, one may apply tools such as SOFAR \citep{SOFAR2019}. For each reference or test text $T_j$ with $0 \leq j \leq m$, denote by $\lambda_{jl}$'s the obtained singular values in decreasing order, and $u_{jl}$'s and $v_{jl}$'s the corresponding $n_j$-dimensional left singular vectors and $p$-dimensional right singular vectors, respectively, with $1 \leq l \leq \min\{n_j, p\}$. Intuitively, an SVD layer with a larger singular value $\lambda_{jl}$ represents a more important latent theme of text $T_j$, while the statistical significances of the left singular vector components in $u_{jl} \in \mathbb{R}^{n_j}$ encode the set of important key words in the associated latent theme (i.e., layer) of text $T_j$.

As mentioned above, 
when the sparse SVD structure is preferred over the regular (nonsparse) SVD structure for the LIDS (see Algorithm \ref{alg:Dir_Met2}), we can resort to 
SOFAR \citep{SOFAR2019} to estimate the latent sparse SVD structure based on the BERT embedding matrix $X_j \in \mathbb R^{n_j\times p}$ corresponding to given text $T_j$ with $j\in \{0, 1,\ldots, m\}$. To this end, we provide a brief review of SOFAR here. Specifically, SOFAR 
assumes the following sparse SVD structure for the BERT embedding matrix 
\begin{equation} \label{new.eq.FL001}
X_j= C_j + E_j= U_j^* \Lambda_j^* V_j^{*\top} + E_j,
\end{equation}
where the deterministic mean matrix $C_j$ admits the sparse singular value decomposition (SVD) 
$C_j = U_j^* \Lambda_j^* V_j^{*\top}$ and $E_j$ is the  noise matrix with independent mean zero entries. 
Here, $U^*_j= \left( u_{j1}^*,\ldots,u_{jk^*}^*\right) \in\mathbb R^{n_j\times k_j^*}$ and $V^*_j = \left( v_{j1}^*,\ldots,v_{jk^*}^*\right) \in\mathbb R^{p\times k_j^*}$ are orthonormal matrices consisting of sparse left and right singular vectors, respectively, $\Lambda^*_j = \text{diag}(\lambda_{j1}^*,\ldots, \lambda_{jk_j^*}^*)$  is the diagonal matrix of nonzero singular values arranged in decreasing order, and $k_j^*$ is the rank of mean matrix $C_j$. 

To estimate the $k_j^*$ sparse SVD layers in $(U_j^*, \Lambda_j^*, V_j^*)$, SOFAR exploits the regularization approach and solves the following minimization problem 
\begin{equation} \label{new.eq.FL002}
(\hat U_j, \hat \Lambda_j, \hat V_j) = {\arg\min}_{U, \Lambda, V} \left\{\frac{1}{2}\|X_j - U\Lambda V\|_F^2  + \lambda_d \|\Lambda\|_1 + \lambda_a(U\Lambda) +  \lambda_b(V\Lambda)\right\}
\end{equation} 
subject to the constraints that $U^TU=I_{k_j}$ and $V^TV=I_{k_j}$, where $\|\cdot\|_F$ denotes the matrix Frobenius norm and $\|\Lambda\|_1$ stands for the sum of the diagonal entries (i.e., the nonnegative singular values). Here, $\lambda_d$, $\lambda_a$, and $\lambda_b$ are nonnegative regularization parameters, and $\rho_a$ and $\rho_b$ are regularization functions used to encourage sparsity in the singular vector matrices. It is important to note that SOFAR estimates all sparse SVD layers simultaneously while preserving the orthogonality constraints. Because of the use of regularization functions, SOFAR estimates are generally biased and not suitable for uncertainty quantification. \cite{zheng2023} proposed a debiasing framework SOFARI for inference based on the SOFAR estimates and established the asymptotic normalities of the resulting debiased estimates that enable the latent SVD inference. We adapt this framework to quantify the uncertainty in LIDS and introduce a corresponding visualization tool in Section \ref{new.Sec.sofarivisual} (see Section \ref{Subsec.SOFARIvis} for the related empirical results).

For each reference or test text $T_j$ with $0 \leq j \leq m$, using the estimated singular values, and left and right singular vectors $(\lambda_{jl},u_{jl},v_{jl})$'s, we can define the associated overall LIDS direction vector $d_j(k) \in \mathbb R^p$ as
\begin{align} \label{def:direction-vector}
    d_j(k) = \sum_{l=1}^k \lambda_{jl}^\alpha s_{jl} \sum_{i=1}^{n_j} u_{jli} w_{ji},
\end{align}
where $k$ is a given positive integer, $\alpha > 0$ is a prespecified constant, $s_{jl} = \text{sign}(\langle v_{jl}, p^{-1/2} 1_p\rangle)$ with $v_{jl}$ the corresponding right singular vector and $1_p \in \mathbb{R}^p$ a vector of ones, $u_{jl} = (u_{jli})_{1 \leq i \leq n_j}$ is the corresponding left singular vector, and $w_{ji} \in \mathbb{R}^p$ is the $i$th row (i.e., token $i$) of the corresponding BERT embedding matrix $X_j \in \mathbb R^{n_j\times p}$. 
Note that the signs $s_{jl}$'s are introduced in (\ref{def:direction-vector}) above to make the overall LIDS direction vector $d_j(k)$ well-defined since the left singular vectors are identifiable only up to a sign change (i.e., $-u_{jl}$ is still a valid left singular vector). Consequently, the left singular vectors $u_{jl}$'s now become the signed version to ensure identifiability. 
Parameter $\alpha > 0$ controls the effect of weighting the latent layers using the singular values, where larger magnitude of a singular value $\lambda_{jl}$ indicates more relative contribution of that specific SVD layer to the variation of text data and hence more importance of the corresponding latent theme in the text. For simplicity, we consider the simple choice of $\alpha = 1$ throughout the paper. 

Intuitively, parameter $k$ (i.e., the number of latent SVD layers utilized) determines the accuracy and quality of the overall LIDS direction vector $d_j(k)$ given in (\ref{def:direction-vector}) for approximating text $T_j$ in the sense that the lower-ranked SVD layers may stem from noise in the text which may deteriorate the similarity between the longer original (i.e., reference) text and the much shorter LLM summary (i.e., test) text. In contrast to the row sum of BERT embedding matrix $X_j$ directly, the benefit of constructing $d_j(k)$ as in \eqref{def:direction-vector} is to eliminate noise in the text and to focus on more significant features of the original text or the LLM summary rather than on the nuanced text details. This is desirable because a good LLM summary will focus primarily on the main points of a text rather than the fine-grained details due to the length constraint. 

In view of (\ref{def:direction-vector}), our definition of $d_j(k)$ is similar to summing over the rows of a low-rank matrix approximation using the SVD. A major difference is that instead of multiplying by the right singular vectors, each a unit vector, we multiply by the corresponding BERT token embedding vectors. It is important to note that the \textit{magnitude} of different BERT token embedding vectors may \textit{differ}. For instance, tokens that appear in a wide range of contexts tend to have their vector representations averaged across those contexts, resulting in vectors with smaller norms that carry less meaning. In contrast, tokens that occur in more specific contexts retain more distinctiveness, leading to embedding vectors with larger magnitude. See, e.g.,  \citet{schakel2015measuringwordsignificanceusing} for details. Thus, the overall LIDS direction vector $d_j(k)$ defined in (\ref{def:direction-vector}) above \textit{emphasizes} tokens with more significant context-specific meanings, while \textit{downgrading} tokens that generally carry less importance.

\begin{algorithm}[tp]
\caption{LIDS: the BERT-SVD-based direction metric and summary embeddings}
\label{alg:Dir_Met2}

\begin{algorithmic}

\State \textbf{Inputs:} Reference text $T_0$ and test texts $\{T_1,\ldots,T_m\}$ with lengths $n_j$ for $j \in \{0,1,\ldots,m\}$.

\State \textbf{for} $j = 0$ \textbf{to} $m$ \textbf{do}

\begin{enumerate}
\item Apply the BERT model to construct the token embedding vectors of dimensionality $p$ for text $T_j$ and obtain an $n_j \times p$ BERT embedding matrix $X_j$.

\item Apply the SVD (or sparse SVD if desired) to the BERT embedding matrix $X_j$ to obtain the empirical singular values $\lambda_{jl}$'s ranked in decreasing order, and the corresponding left singular vectors $u_{jl}$'s and right singular vectors $v_{jl}$'s for $1 \le l \le \min\{n_j,p\}$.

\item Calculate the overall LIDS direction vectors $d_j(k) \in \mathbb{R}^p$ defined in \eqref{def:direction-vector} for $1 \le k \le \min\{n_j,p\}$ for text $T_j$.
\end{enumerate}

\State \textbf{end for}

\State \textbf{for} $j = 1$ \textbf{to} $m$ \textbf{do}

\begin{enumerate}
\item Calculate the maximum absolute cosine similarity $\text{MACS}_j$ defined in \eqref{def:CosineSimilarity} between test text $T_j$ and reference text $T_0$.

\item Obtain $\widehat{k}_j \in \{1, \cdots, \min\{n_j, p\}\}$ that maximizes $\left|\text{CS}(d_j(k), d_0(k))\right|$ with respect to $k$.
\end{enumerate}

\State \textbf{end for}

\State \textbf{return} The LIDS similarity measures $\text{MACS}_j$'s and the associated $p$-dimensional LIDS summary embeddings $d_j(\widehat{k}_j)$ for $1 \le j \le m$.

\end{algorithmic}
\end{algorithm}

For each given pair of shorter test (i.e., LLM summary) text and longer reference (i.e., original) text $(T_j, T_0)$ with $1 \leq j \leq m$, we are now ready to define the latent BERT-SVD-based LIDS direction metric as 
\begin{align} \label{def:CosineSimilarity}
                \text{MACS}_j = \max_{1 \leq k \leq \min\{n_j, p\}} \left|\text{CS}(d_j(k), d_0(k))\right|,
            \end{align}
where $d_j(k)$ and $d_0(k)$ are given in (\ref{def:direction-vector}), $\text{CS}(d_j(k),d_0(k)) = \langle d_j(k), d_0(k)\rangle/(\|d_j(k)\|_2 \|d_0(k)\|_2)$ denotes the cosine similarity between two vectors with $\|\cdot\|_2$ the vector Euclidean norm, and $n_j \leq n_0$ by assumption. Since the LLM summary is usually much shorter than the original text, it is sensible to measure their similarity through pairs of their overall LIDS director vectors with aligned numbers of latent SVD layers. If a summary is accurate and of high quality, it is naturally expected that the summary will capture the most important themes of the original text that are encoded in the leading latent SVD layers, where the remaining SVD layers account for less important text details that could not be retained due to the length constraint. 

Our latent BERT-SVD-based LIDS direction metric introduced in (\ref{def:CosineSimilarity}) above provides a new similarity measure between an LLM summary $T_j$ and the original text $T_0$ by maximizing such paired cosine similarity in magnitude over the number of latent layers $k$ for text reduction and approximation. In particular, in contrast to existing approaches, LIDS utilizes layered pairs of the singular values and singular vectors weighted tokens characterizing the underlying themes in texts with the latent SVD layers. See Sections \ref{new.sec.relawork} and \ref{new.sec.ourinno} for detailed discussions on the key differences between  the LIDS similarity
measure and the BERTScore \citep{Zhangetal2019BERTScore} as well as other related approaches. In light of (\ref{def:CosineSimilarity}), the LIDS similarity measure ranges between $0$ and $1$, where $0$ indicates that the summary and original text have nothing in common, and $1$ indicates that the summary and original text are essentially identical (with possibly different wordings). Algorithm \ref{alg:Dir_Met2} summarizes the major steps for the first part of the LIDS framework.

\subsection{LIDS summary embeddings} \label{new.sec.lids.summ.embed}

For down-stream text applications, it is often desirable to generate embeddings of the LLM summaries for large text reduction. For each given LLM summary $T_j$, maximizing the absolute cosine similarity between pairs of the overall LIDS director
vectors of the summary and the original text $T_0$ with aligned numbers of latent SVD layers over parameter $k$ in (\ref{def:CosineSimilarity}) gives the optimal choice of $k = \widehat{k}_j \in \{1, \cdots, \min\{n_j, p\}\}$. Then the associated overall LIDS director
vector $d_j(\widehat{k}_j)$ in (\ref{def:direction-vector}) provides a natural $p$-dimensional embedding of summary text $T_j$, as summarized in Algorithm \ref{alg:Dir_Met2}. We will refer to $d_j(\widehat{k}_j) \in \mathbb{R}^p$ as the LIDS summary embedding throughout the paper. Such text-level LIDS embeddings are more concise and holistic than the individual token-level BERT embeddings, serving as a useful tool for large text reduction. Further insights into those $\widehat{k}_j$ latent SVD layers will be provided in Section \ref{new.Sec.sofarivisual} through layered key word selection with controlled error rate.

\subsection{Two benchmark summary mechanisms} \label{new.sec.twobench}

Ideally, the LIDS similarity measure in (\ref{def:CosineSimilarity}) should possess the capability of distinguishing between different levels of accuracy and quality of the summary texts relative to the original text. To help validate such intuition, we will introduce two benchmark summary mechanisms for large text reduction. The first one is the naive summary and the second one is the random topic summary, as described below.

\textit{Naive summary}. Since we analyze words and tokens as vectors, the objective of summary for text reduction is to approximate the overall direction vector of a large collection of vectors with that of a smaller collection of vectors. A simple 
way to approximate the overall direction vector with fewer vectors is to sample from a large collection of vectors, giving rise to the first benchmark of naive summary. The naive summary simply samples words from the original text $T_0$. Thus, the more frequently the word occurs in the original text, the more likely that word will be chosen for the random sample. 
The naive summary is by no means a real summary since it has no punctuation or meaning in the orderings of the words. 
For a fair comparison, the number of words used in each naive summary matches that used in each corresponding LLM summary. Given that word orderings are completely removed, it is natural to expect that the naive summary will be \textit{much less accurate} than the LLM summary when the BERT token embeddings are considered.

\textit{Random topic summary}. The second benchmark is the random topic summary. We get $50$ different subjects such as Quantum Mechanics, Plato's Republic, The French Revolution, and Leonardo Da Vinci by asking ChatGPT for varying topics. We utilize ChatGPT again to generate a summary about each topic. We then verify that each summary is coherent on the desired topic. The random topic summaries are of similar lengths to summaries generated from the original text $T_0$. Intuitively, the random topic summary will be \textit{less similar} to original text $T_0$ compared to the LLM summary since random topics are unrelated to the original text.

\section{FDR control for layered key word selection with SOFARI} \label{new.Sec.sofarivisual}

To gain some useful insights into the $\widehat{k}_j$ latent SVD layers associated with the LIDS summary embedding vector $d_j(\widehat{k}_j)$ given in Section \ref{new.sec.lids.summ.embed} for text $T_j$, in this section we present the second part of the LIDS framework on the layered key word selection with controlled error rate via SOFARI. The major goal is to visualize important key words underlying main themes of the LLM summary with inference guarantees.

For the implementation of the second part of the LIDS framework (i.e., LIDS visualization), 
we need to construct valid p-values for components of left singular vectors $u_{jl}$'s with $1 \leq l \leq k_j$ from the corresponding BERT embedding matrix $X_j \in \mathbb R^{n_j\times p}$ so that we can apply the BH procedure for FDR control toward layered key word selection with inference guarantees. In particular, the recently developed inference tool of SOFARI \citep{zheng2023} provides uncertainty quantification on SOFAR estimates through theory-guided p-values and confidence intervals on the leading singular values and corresponding left singular vectors. By leveraging the ideas of Neyman near-orthogonality condition on the score function \citep{Chernozhukov2018} and manifold-based inference, SOFARI provides bias correction on the SOFAR estimates, 
based on which justified p-values and confidence intervals are constructed to quantify the uncertainty of these estimates for the singular values and left singular vectors. 
See \citet{SOFAR2019} and \citet{zheng2023} for details on the implementation and theoretical properties of SOFAR and SOFARI methods, respectively.

For each given LLM summary $T_j$, we have the estimated left singular vectors $u_{jl}$'s from the corresponding BERT embedding matrix $X_j \in \mathbb R^{n_j\times p}$. Observe that for each $1 \leq l \leq \min\{n_j, p\}$, the left singular vector $u_{jl}$ is $n_j$-dimensional and its components represent the compositions and weights of tokens in the $l$th SVD layer, encoding the importance level of each token in that specific layer. As mentioned above, we exploit the latent SVD inference tool of SOFARI \citep{zheng2023} to calculate the p-values for tokens corresponding to the components of the left singular vector $u_{jl}$. We then apply the Benjamini--Hochberg (BH) procedure \citep{benjamini1995controlling} to identify and visualize a set of important key words for the $l$th layered theme with the false discovery rate (FDR) controlled at a target level $q \in (0, 1)$, 
meaning that on average, the faction of falsely selected key words is under control at level $q$. Such visualization (e.g., through word cloud plots) depicts the major latent themes (in ranked importance by the singular values) underlying the LLM summary with statistically significant key words. The size of each selected key word in a latent SVD layer is determined by the magnitude of the corresponding SOFARI test statistic for that specific left singular vector component.

When varying the desired length of the LLM summary in the prompt, we may obtain a \textit{zoomed-out} or \textit{zoomed-in} view of the latent themes that are key to the original large text. As such, LIDS gives a principled, quantified way of large text reduction. In contrast to existing approaches, the visualization in LIDS enables us to gain useful insights into large text reduction with statistical guarantees for effective LLM summary inference.

\section{Empirical studies} \label{new.Sec.empistud}

In this section, we aim to provide comprehensive evaluations of LIDS for LLM summary inference through six different perspectives: 1) two benchmark summary mechanisms, 2) human verification, 3) comparisons to other similarity metrics, 4) LIDS visualization for summary inference, 5) comparison of different LLMs, and 6) robustness checks across varying text domains in Sections \ref{new.Sec.gptsumminte}--\ref{Subsec.DocDomains}, respectively, via text-based empirical studies.

\subsection{LIDS validation through two benchmark summary mechanisms} \label{new.Sec.gptsumminte}

To demonstrate the main points, we will focus on a Utah article for now and further investigate three additional text applications later in Section \ref{Subsec.DocDomains} (i.e., a NASA article, a legal document, and a novel chapter). The Utah article is of $1301$ words and about a Utah family who bought a home and is suing the previous owner over mold issues, while the previous owner is also being charged with murdering her husband; see \url{https://www.nbcnews.com/news/us-news/kouri-richins-utah-family-sues-house-mold-update-rcna111488}. 
To evaluate the performance of LIDS, we focus on the recent release of GPT-5 for generating the LLM summary of the Utah article. To obtain an empirical distribution for the LIDS similarity measure in  \eqref{def:CosineSimilarity} between the LLM summary and the original article, we repeat the prompt of summary with desired length $50$ times. The two benchmark summary mechanisms introduced in Section \ref{new.sec.twobench}, the naive summary and the random topic summary, are also implemented with $50$ random repetitions. As a result, we obtain the empirical distributions for all three summary methods. 

\begin{figure}[t]
    \centering
    \includegraphics[width=0.60\textwidth]{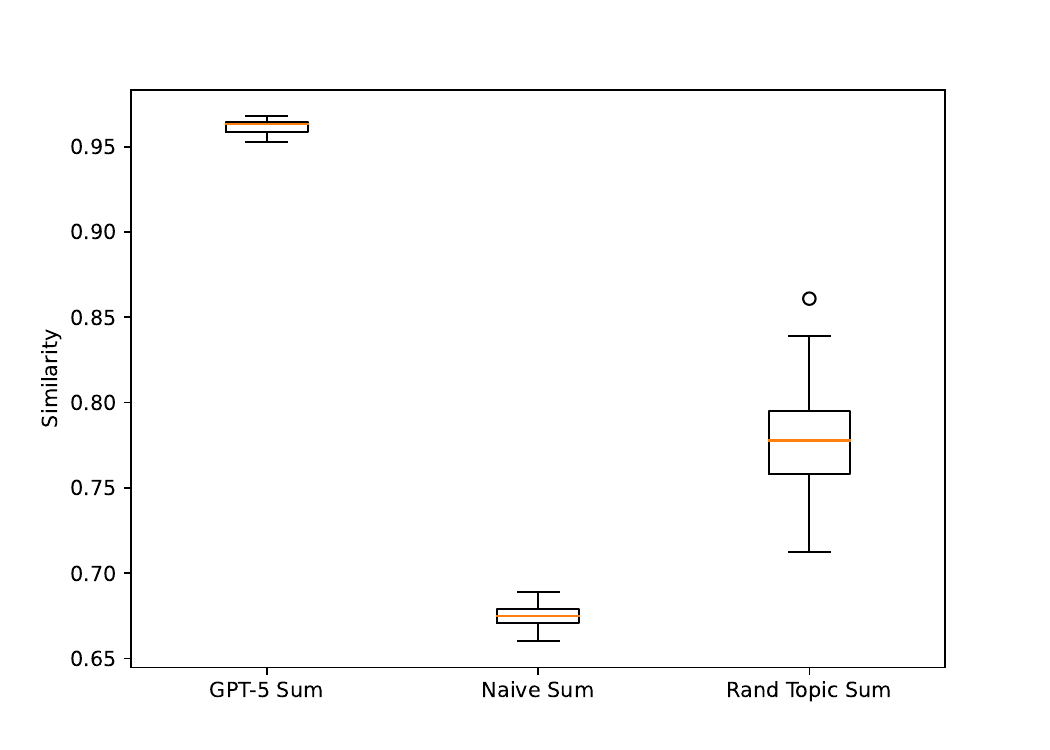}
    \caption{Boxplots of the LIDS similarity measures for GPT-5 over $50$ repeated prompts and two benchmark summary mechanisms over $50$ random repetitions on the Utah article.
    }
    \label{fig:GPT_Verification_Utah}
\end{figure}

From Figure \ref{fig:GPT_Verification_Utah} as well as Table \ref{table:GPT_Verification_Utah} in Section \ref{Subsec.horizcomp.supp} of the Supplementary Material, it can be seen that the GPT-5 summary performs the best under the LIDS similarity measure. In particular, the similarity measures of the naive summary and random topic summary do \textit{not} overlap with those of the GPT-5 summary as the lowest score of the GPT-5 summary is approximately $0.95$, whereas the highest scores of the naive and random topic summaries are approximately $0.69$ and $0.87$, respectively. As such, LIDS gives a clear differentiation between the accuracy and quality of these three summary methods, with the GPT-5 summary exhibiting \textit{much higher} LIDS similarity measure and \textit{much lower} variance compared to the two benchmark summary mechanisms. It is intuitive to expect the relatively poor performance of the random topic summary since it is in an unrelated topic although the writing itself is smooth. In contrast, the naive summary performs consistently worse mainly because the orderings of words and tokens are now completely removed, resulting in much deviated BERT token embedding vectors. These results support that the LLM summary (e.g., ChatGPT) is indeed intelligent in accurately and stably reducing and summarizing large texts.

\subsection{LIDS validation through human verification} \label{Subsec.HumanVerif}

Although time-consuming, human evaluation is the golden standard for assessing the accuracy and quality of text summaries. To verify the validity of LIDS, we design an experiment to collect human evaluation data. 
Using the Utah article, we create $30$ summaries of varying quality. The goal is to determine whether the LIDS similarity measure can differentiate between excellent, good, satisfactory, poor, and very poor summaries of the Utah article in the same way that humans can. To guide the human evaluation, we create a rubric with the aid of an English language expert. The use of an English language expert is to avoid bias that would align the rubric scoring with how the LIDS similarity measure operates. The summary evaluation rubric employs four criteria: 1) coverage of ideas in original text, 2) objective retelling of original text, 3) grammatical organization and clarity, and 4) brevity and focus. All criteria are weighted equally and evaluated with points from $1$--$5$, where $1$ means the worst and $5$ indicates the best. See Section \ref{Subsec.HumanVerif.supp} of the Supplementary Material for the full instructions given to the participants and the rubric used in this experiment. 

\begin{figure}[t]
    \centering
    \includegraphics[width=0.7\textwidth]{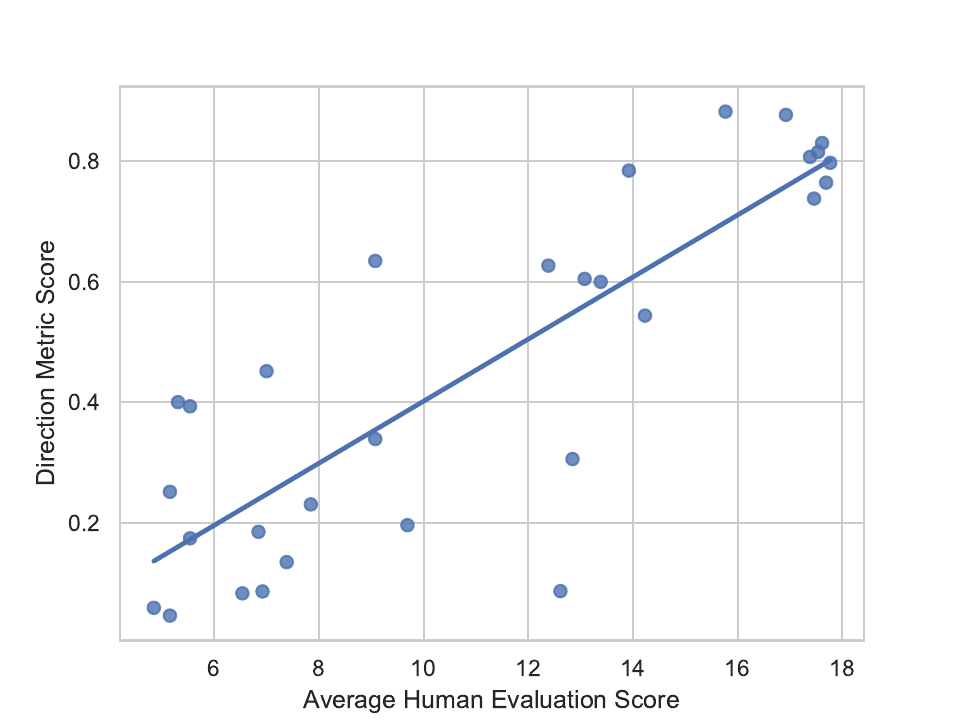}
    \caption{Scatter plot of the average human evaluated summary quality scores on the horizontal axis and the LIDS similarity measure scores of the summaries on the vertical axis, with a linear regression line showing a Pearson correlation of $0.904$
    between the two, for the experiment in Section \ref{Subsec.HumanVerif}.}
    \label{fig:Human_Corr}
\end{figure}

We have a set of $48$ participants conducting independent human evaluations. For each of the $30$ summaries generated above, we average the scores from those $48$ participants to obtain an average overall score. Taking the average over all the participants scores can help alleviate the individual biases and variations that occur. As a result, we obtain a total of $30$ aggregated human evaluation data points, in which each data point represents an average score of $48$ participants for the corresponding summary. We also calculate the values of the LIDS similarity measure for those $30$ summaries. Figure \ref{fig:Human_Corr} depicts a scatter plot between the LIDS similarity measure and the human evaluation. Indeed, there is a \textit{strong} linear correlation of $0.904$ with a $95\%$ confidence interval of $[0.81, 0.95]$ between them, confirming that the LIDS similarity measure is able to distinguish effectively the quality of different summaries in a similar manner to how humans can. 

As a robustness check, we also examine the distance correlation (a nonlinear correlation) and Kendall's tau correlation (a robust correlation) between the LIDS similarity measure and human evaluations. LIDS achieves a distance correlation of $0.873$ with a $95\%$ confidence interval of $[0.77, 0.95]$ and a Kendall's tau correlation of $0.664$ with a $95\%$ confidence interval of $[0.51, 0.79]$. We see that the LIDS similarity measure correlates strongly with human evaluations across all three forms of correlation.

\subsection{LIDS validation through comparisons to other similarity metrics} \label{Subsec.SimMetVal}

We further verify the efficacy of the LIDS similarity measure by comparisons to some popular similarity metrics such as the BLEU, ROUGE-1, ROUGE-L, METEOR, and BERTScore. 
Figure \ref{fig:SimMets_Vs_Bench_Utah} in the Introduction depicts the performances of the LIDS, BLEU, ROUGE-1, ROUGE-L, METEOR, and BERTScore, respectively, in comparing the LLM summary (e.g., GPT-5) to the naive summary and random topic summary on the Utah article, using the rescaled boxplots for a meaningful visual comparison. 
From Figure \ref{fig:SimMets_Vs_Bench_Utah}, 
we see that most similarity metrics perform similarly in pattern to the LIDS direction metric in that those metrics correctly categorize that the LLM summary (e.g., ChatGPT) outperforms the two benchmark summary mechanisms, but there is generally overlap between the boxplots for the LLM summary and the two benchmark summary mechanisms. This is not true across the board as ROUGE-1 fails to pass the naive summary benchmark. BERTScore also exhibits strong performance by definitively classifying the LLM summary and the two benchmark summary mechanisms with no overlap similarly to LIDS, with LIDS outperforming BERTScore. Such results give credibility to both the LIDS direction metric and the two benchmark summary mechanisms as the benchmarks are not trivially easy for all similarity metrics to pass. 

\begin{table}[tp]
\centering
\caption{Computational costs of different similarity metrics over $50$ summaries on the Utah article
}
    \begin{tabular}{lc}
        \hline
        Similarity Metric & Seconds \\
        \hline
        LIDS & 25.519812 \\
        BLEU & 0.629676 \\
        ROUGE-1 & 8.013586 \\
        ROUGE-L & 32.912094 \\ 
        METEOR & 8.805413 \\
        BERTScore & 158.460620 \\
        \hline
    \end{tabular}
    \label{table:Comp_Time_50_Sums}
\end{table}

\begin{table}[htp]
 \centering
\caption{Memory usages of different similarity metrics over $50$ summaries on the Utah article
}
    \begin{tabular}{lcc}
        \hline
       Similarity Metric & \multicolumn{2}{c}{Memory Usage (MB)}  \\
        \cline{2-3}
         & Incremental & Peak Usage \\
        \hline
        LIDS & 19.65 & 30.04 \\
        BLEU & 1.87 & 2.08 \\
        ROUGE-1 & 0.05 & 0.20 \\
        ROUGE-L & 0.03 & 2.78 \\
        METEOR & 86.37 & 113.99 \\
        BERTScore & 10.17 & 48.36 \\
        \hline
    \end{tabular}
    \label{table:Mem_Use}
\end{table}

\begin{table}[htp]
\centering
\caption{The linear (Pearson) correlations, distance correlations, Kendall's tau correlations, and the associated 95\% confidence intervals (CIs) between different similarity metrics and human evaluations for the experiment in Section \ref{Subsec.HumanVerif}
}
    \begin{tabular}{l|cc|cc|cc} 
        \hline
       Similarity Metric 
       & \multicolumn{2}{|c|}{Pearson Cor 
       } 
       & \multicolumn{2}{|c|}{Distance Cor 
       } 
       & \multicolumn{2}{|c}{Kendall's Tau 
       }\\
       \hline
        LIDS & $0.904$ & $[0.81, 0.95]$ & $0.873$ & $[0.77, 0.95]$ & $0.664$ & $[0.51, 0.79]$\\
        BLEU & $0.232$  & $[-0.14, 0.55]$ & $0.135$ & $[0.00, 0.37]$ & $0.370$ & $[ 0.08, 0.60]$ \\
        ROUGE-1 & $0.685$  & $[0.43, 0.84]$ & $0.567$ &  $[0.35, 0.75]$ & $0.568$ & $[0.40, 0.70]$ \\ 
        ROUGE-L & $0.733$  & $[0.51, 0.86]$ & $0.604$ &  $[ 0.38, 0.77]$ & $0.575$ &  $[0.37, 0.72]$ \\
        METEOR & $0.662$  & $[0.40, 0.83]$ & $0.575$ &  $[0.35, 0.76]$ & $0.598$ &  $[0.43, 0.74]$ \\
        BERTScore & $0.976$  & $[0.95, 0.99]$ & $0.955$ &  $[0.93, 0.98]$ & $0.880$ &  $[0.79, 0.95]$ \\
        \hline
    \end{tabular} 
    \label{table:LLM_Human_Comp}
\end{table}

We report in Tables \ref{table:Comp_Time_50_Sums} and \ref{table:Mem_Use} comparisons of different similarity metrics in terms of the computational cost and memory usage, respectively, over $50$ summaries on the Utah article. In particular, in view of Table \ref{table:Comp_Time_50_Sums}, it can be seen that LIDS is computationally more efficient than both BERTScore and ROUGE-L. From Table \ref{table:Mem_Use}, it is evident that LIDS is more efficient than both BERTScore and METEOR in terms of peak usage of memory. 

We further examine in Table \ref{table:LLM_Human_Comp} the linear (Pearson) correlations, distance correlations (a nonlinear correlation), and Kendall's tau correlations (a robust correlation) as well as the corresponding 95\% confidence intervals (CIs) between different similarity metrics and human evaluations for the experiment in Section \ref{Subsec.HumanVerif}. From Table \ref{table:LLM_Human_Comp}, we see that both LIDS and BERTScore outperform the other similarity metrics, while the 95\% CIs of LIDS and BERTScore overlap with each other across different forms of correlation.

In summary, both the LIDS similarity measure and BERTScore significantly outperform the BLEU, ROUGE-1, ROUGE-L, and METEOR in distinguishing between the LLM summary and the two benchmark summary mechanisms as well as in terms of different correlations with human evaluations, while LIDS is \textit{computationally more efficient} than BERTScore.

\begin{figure}[t]
    \centering
    
    \includegraphics[trim={0.85in 0 0 0},clip, width=\textwidth
    ]{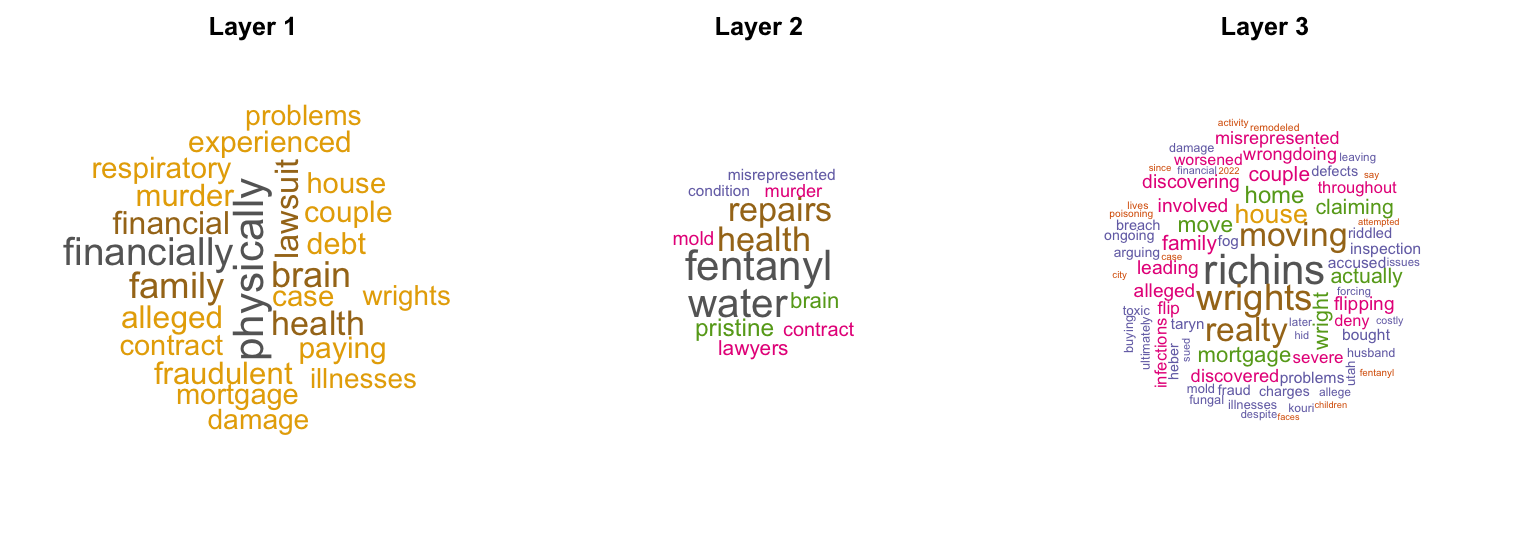}
    \caption{LIDS visualization word cloud plots with FDR control at level $q = 0.005$ for the first three latent SVD layers of a representative LLM summary of the Utah article.}
    \label{fig:Utah_Clouds}
\end{figure}

\subsection{LIDS visualization for LLM summary inference} \label{Subsec.SOFARIvis}

Given the efficacy of the LIDS similarity measure, it is practically valuable to interpret the mechanisms underlying high-quality LLM summaries (e.g., ChatGPT). We explore this through the LIDS visualization framework proposed in Section \ref{new.Sec.sofarivisual}, which facilitates refined inference of these summaries. Section \ref{Subsec.SOFARIvis.supp} of the Supplementary Material provides the full details on how LIDS enables us to decompose the LLM summary into layered word cloud plots of ranked importance. Such plots clearly show important key words underlying the latent themes of the summary. 
A larger size of a selected key word indicates higher statistical significance in the corresponding latent SVD layer (i.e., that specific left singular vector). Figure \ref{fig:Utah_Clouds} presents the LIDS visualization word cloud plots associated with the first three latent SVD layers of a representative LLM summary of the Utah article, where the FDR is controlled at level $q = 0.005$ and the chosen GPT-5 summary has $\widehat{k}_j \geq 3$ from Algorithm \ref{alg:Dir_Met2}. 
Note that the BERT embeddings we work with are not for individual words, but are for specific tokens. Some of these tokens are punctuation marks or even partitions of individual words due to the way that BERT performs its embeddings. To deal with such issue, we recover the original words that were split up by recombining the corresponding tokens, and the size of such word in the word cloud plot is determined by the maximum magnitude of the corresponding SOFARI test statistics for those tokens. We further eliminate stop words and punctuation since they do not provide meaningful information to the word cloud plots to decipher the meaning of the text.

From the layer $1$ word cloud plot in Figure \ref{fig:Utah_Clouds}, we see indications of a lawsuit that impacts a family both physically and financially. The presence of terms related to a house and mortgage suggests real-estate issues, and there are even hints of a possible murder. The layer $2$ word cloud plot in Figure \ref{fig:Utah_Clouds} reinforces the idea of significant legal and financial repercussions, while also highlighting the role of mold in the dispute.
In the layer $3$ word cloud plot in Figure \ref{fig:Utah_Clouds}, names of those involved begin to appear, and it becomes evident that the situation involved an unexpected discovery about the home—likely one that had been flipped—given the recurrence of the words flip and flipping in this layer and earlier in the layer $1$ word cloud plot. 
As such, the LIDS visualization with FDR control enables us to gain statistically justified intuitions on the overarching text in the LLM summary of the Utah article. This is important not just so that we as humans can understand the text, but so that we can understand the underlying work done by LLMs in what makes these summaries intelligent. As a result, we can dissect the most important themes uncovered by the LLM in crafting a good LLM summary of the original text. 

\begin{figure}[t]
    \centering
    \includegraphics[width=0.90\textwidth]{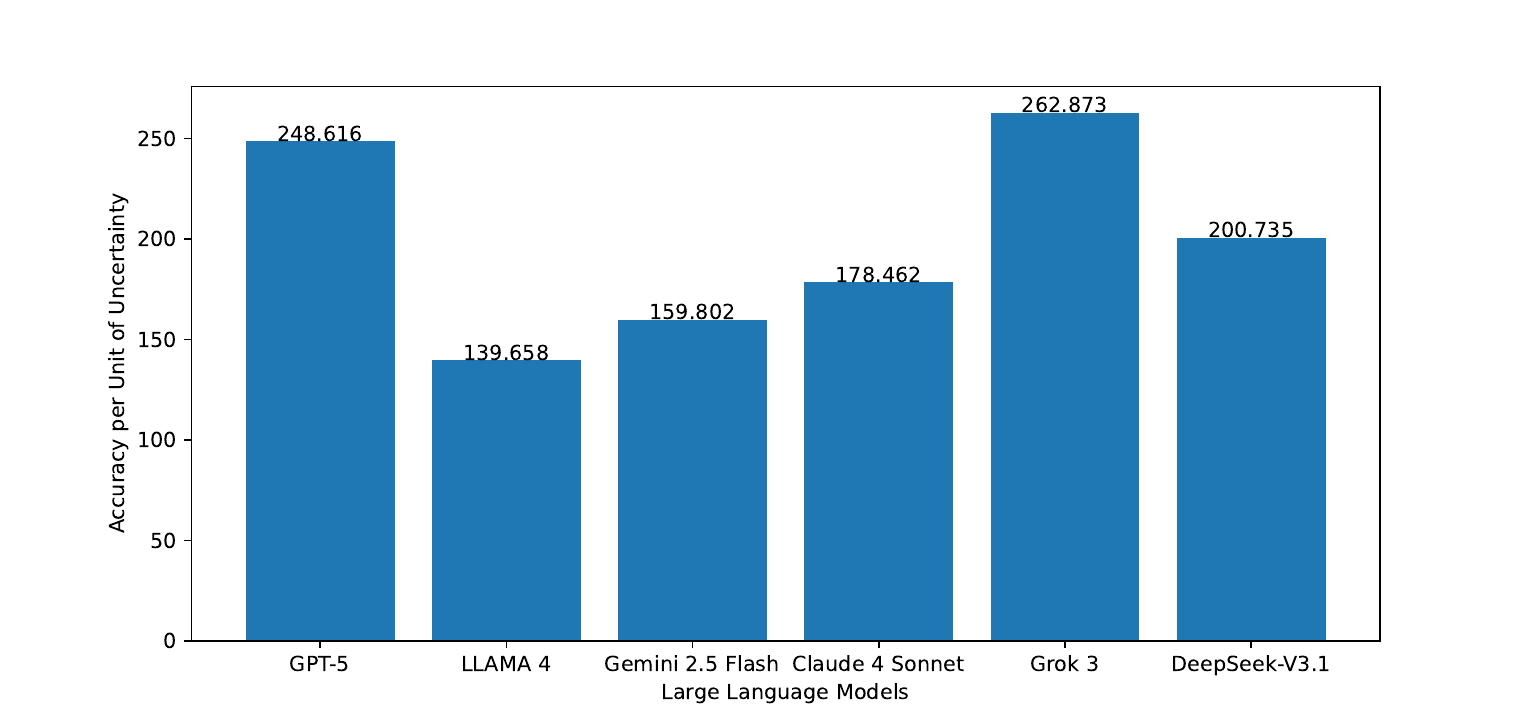}
    \caption{
    Comparison of different LLMs with LIDS in terms of the Sharpe ratio-type measure of accuracy per unit of uncertainty, i.e., the mean similarity divided by the corresponding standard deviation over $50$ repeated prompts on the Utah article. Larger values indicate better performance. 
    }
    \label{fig:LLM_Sharpe_Comp_Utah} 
\end{figure}

\subsection{Comparison of different LLMs} \label{Subsec.horizcomp}

We also conduct a comparison of different LLMs: ChatGPT, Claude, DeepSeek, Gemini, Grok, and Llama, based on the LIDS similarity measure. 
Since each LLM gives an empirical distribution of the similarity measure relative to the original article through $50$ repeated prompts, one natural idea is to exploit the Sharpe ratio-type measure \citep{Sharpe1966}, 
which is defined as the mean divided by the corresponding standard deviation (SD) of the distribution. 
The results are depicted in Figure \ref{fig:LLM_Sharpe_Comp_Utah}, which enables us to see clearly how effective and robust each LLM is at summarizing large text across repeated prompts, with higher values indicating better performance. 

Overall, all the LLMs fare well on summarizing the Utah article under the lens of LIDS, with GPT-5 and Grok 3 at the top in terms of the Sharpe ratio-type measure of accuracy per unit of uncertainty (i.e., the mean similarity divided by the corresponding
standard deviation), as seen in Figure \ref{fig:LLM_Sharpe_Comp_Utah}. 
See Section \ref{Subsec.horizcomp.supp} of the Supplementary Material for the detailed results.

\subsection{Robustness checks across varying text domains} \label{Subsec.DocDomains}

We finally perform the robustness checks of LIDS over different text domains. To this end, let us further consider a NASA article. The NASA article consists of $1601$ words and is about the landing of NASA's Curiosity rover on Mars in 2012; see \url{https://www.space.com/16932-mars-rover-curiosity-landing-success.html}. The detailed results of LIDS on the NASA article are included in Section \ref{new.sec.nasa} of the Supplementary Material, echoing LIDS's efficacy for LLM summary inference.

\begin{figure}[t]
    \centering
    \includegraphics[trim={0.85in 0 0 0},clip, width=\textwidth]{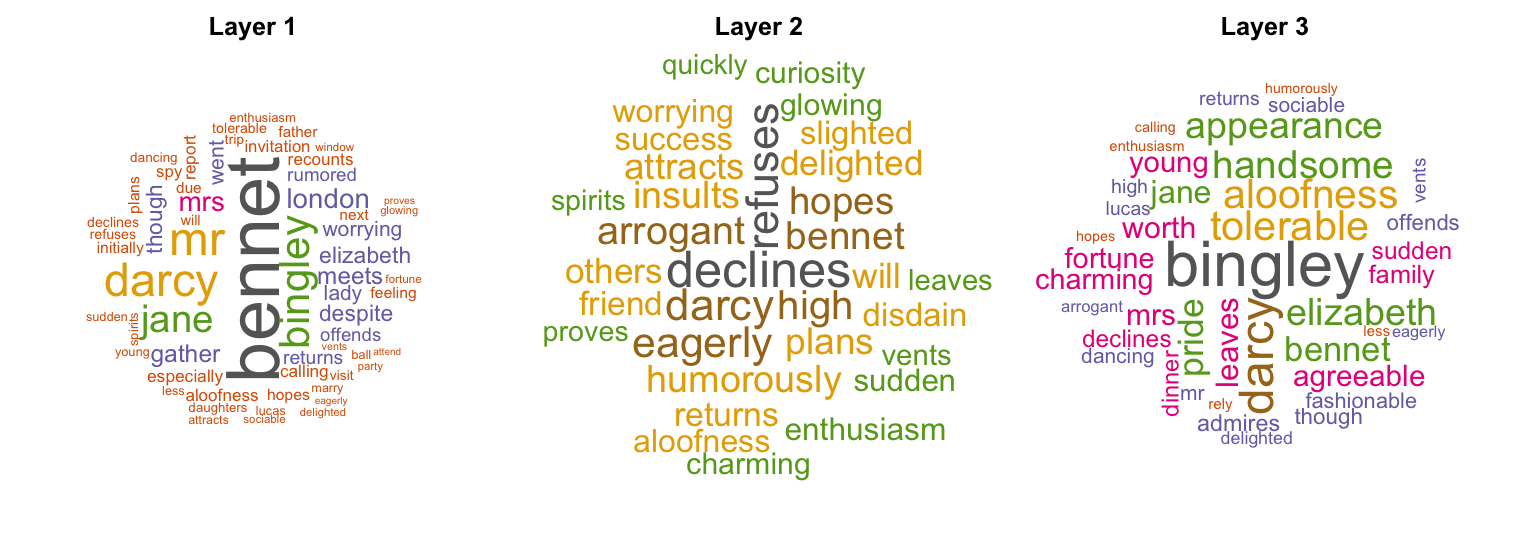}
    \caption{LIDS visualization word cloud plots with FDR control at level $q = 0.005$ for the first three latent SVD layers of a representative LLM summary of the novel chapter (i.e., Chapter $3$ of ``Pride and Prejudice'').}
    \label{fig:PridePrej_Clouds}
\end{figure}

We further examine the performance of LIDS in two additional text domains: a legal document and a chapter from a classic novel, as these are written differently than the Utah and NASA articles. The legal documents are almost always carefully crafted in language and each word used is important. As such, summarizing legal documents can be generally challenging (e.g., for written contracts). The legal document we consider is a school district's alcohol and drug free policy from 2022, available at \url{https://kb.nebo.edu/sites/default/files/2025-01/GBCC%20-%20Alcohol%20and%20Drug%20Free%20Workplace_2022-02-09.pdf}. The novel chapter that we consider is Chapter $3$ of ``Pride and Prejudice,'' a classic novel by Jane Austen.  
Figures \ref{fig:Legal_Domain} and \ref{fig:Story_Domain} in Sections \ref{new.sec.legal} and \ref{new.sec.novel} of the Supplementary Material, respectively, present the results from these two additional text domains. We continue to see that LIDS performs well in revealing that the LLM summary (e.g., ChatGPT) excels over the two benchmark summary mechanisms, affirming the practical utility and robustness of LIDS for LLM summary inference across different text domains.

Moreover, Figure \ref{fig:Legal_Clouds} in Section \ref{new.sec.novel} and Figure \ref{fig:PridePrej_Clouds} depict the LIDS visualization word cloud plots associated with the first three latent SVD layers of a representative LLM summary for each of the two additional text domains, respectively, with FDR control at level $q = 0.005$. 
From the layer $1$ word cloud plot in Figure \ref{fig:PridePrej_Clouds}, we observe that the most prominent names are Bennet, Bingley, Darcy, and Jane. The scene appears to involve a social gathering where people meet, rumors circulate, and someone ultimately takes offense. The setting may be London.
In the layer $2$ word cloud plot in Figure \ref{fig:PridePrej_Clouds}, it becomes evident that both highly negative and highly positive events occur involving Bennet and/or Darcy. Negative terms such as refuses, slighted, insults, disdain, and arrogant appear alongside positive terms like attracts, success, delighted, eagerly, humorously, and charming.
The layer $3$ word cloud plot in Figure \ref{fig:PridePrej_Clouds} again highlights these same central names but adds more descriptive language, including tolerable, aloofness, handsome, pride, charming, agreeable, and fortune.

\section{Discussions} \label{new.Sec.discuss}

We have investigated in this paper the problem of LLM summary inference. In contrast to existing approaches to summary text evaluation, our suggested LIDS framework employs the BERT-SVD-based direction metric using the BERT model for token embeddings. The use of the latent SVD structure for weighting the tokens through both the singular values and singular vector components provides a layered view for summary inference, characterizing the accuracy and uncertainty of the LIDS similarity measure between the original text and the LLM summary. It naturally gives rise to the overall LIDS summary embeddings for large text reduction that be useful for downstream text applications. The layered lens of LIDS is further empowered by the FDR control with SOFARI for unveiling important key words underlying each latent theme of the summary, ensuring statistical inference guarantees.

LIDS is currently based on the BERT embeddings of tokens. It would be interesting to consider more general embedding models such as time series extensions of BERT for constructing more flexible, accurate LIDS direction metric. Also, it would be useful to incorporate the network idea of graph neural networks (GNNs) exploiting the text knowledge graph \citep{Scarsellietal2009,Micheli2009,Wuetalbook2022}. Another interesting problem is how to design ensemble LLM summary inference based on a trajectory of prompts from a single or multiple LLMs. These problems are beyond the scope of the current paper and will be interesting topics for future research.

\section*{Data availability statement}
The data that supports the findings of this study is available from the corresponding author, DP, upon reasonable request.

\bibliographystyle{chicago}
\bibliography{references}

\newpage
\appendix
\setcounter{page}{1}
\setcounter{section}{0}
\setcounter{equation}{0}
\renewcommand{\theequation}{A.\arabic{equation}}

\begin{center}{\bf \Large Supplementary Material to ``LIDS: LLM Summary Inference Under the Layered Lens''}

\bigskip

Dylan Park, Yingying Fan and Jinchi Lv
\end{center}

\noindent This Supplementary Material contains the additional details of the implementation for LIDS and additional empirical results.

\section{Additional details for the LIDS implementation} \label{sec.supp.A}

To generate the LLM summaries, we use a generic LLM prompt ``\texttt{Summarize the following text.}'' preceding a reference text of interest (i.e., original text). The length of the LLM summary was not an issue, with the GPT-5 generated summaries having about $100$ to $200$ words in length. To ensure a set of randomly generated LLM summaries (as opposed to identical ones), we delete past prompts for each new LLM summary. 

All the empirical experiments in this paper have been run on a 2017 Macbook Pro with a 2.2 GHz 6-core intel i7 processor and Radeon Pro 555X 4 GB Intel UHD Graphics 630 1536 MB graphics card. In particular, those numerical studies are \textit{not} computationally expensive with the aid of the Python and NLP packages mentioned in the main body of the paper.

\section{Additional empirical results} \label{sec.supp.B}

In Sections \ref{Subsec.HumanVerif.supp}--\ref{Subsec.DocDomains.supp} below, we present additional empirical details and results for the numerical studies in Sections \ref{Subsec.HumanVerif} and \ref{Subsec.SOFARIvis}--\ref{Subsec.DocDomains}, respectively.

\subsection{LIDS validation through human verification} \label{Subsec.HumanVerif.supp}

\textit{Human verification instructions and rubric}. The instructions given to the set of $48$ participants are as follows. \\
Please follow the instructions below: 
\begin{enumerate}
    \item[1)] Open the following link, which contains all required study materials: ``drive link''.
    \item[2)] Read the evaluation rubric (``Summary\_Evaluation\_Rubric.pdf")
    \item[3)] Read the full news article (``Full\_Article.pdf" or at \\
https://www.nbcnews.com/news/us-news/kouri-richins-utah-family-sues-house-mold-update-rcna111488).
    \item[4)] Review the sample evaluations (``Summary\_Evaluation\_Examples.pdf''). 
    \item[5)] Download  ``Summary\_Evaluations.xlsx" using Excel or Google Sheets.
    \item[6)] Read and evaluate each summary found in ``Summaries.pdf'' using the four rubric criteria.
    \item[7)] Email your completed spreadsheet to me at dylanpar@usc.edu, or share your Google Sheets file with the same email address.
\end{enumerate}

The rubric scores over four categories: 1) coverage of ideas in original text, 2) objective retelling of original text, 3) grammatical organization and clarity, and 4) brevity and focus. These four categories each with the scale of 1--5 give a total summary evaluation score ranging from 4--20. 

\textit{Summary evaluation rubric:} \\ 
1). Coverage of ideas in original text (scale of 1--5) \\
Captures all central ideas and important points from the original text.

\begin{etaremune}
  \item Excellent: All key ideas and important details are accurately and completely included. 
  \item Good: Most key ideas are included with minor omissions. 
  \item Satisfactory: Some key ideas are captured, but several important points are missing. 
  \item Poor: Few key ideas are included; many important points are missing. 
  \item Very Poor: The summary misses all central ideas and maybe contains unimportant ideas.
\end{etaremune} 

\noindent
2). Objective retelling of original text (scale of 1--5) \\
Represents the original text accurately without distortion, bias, or personal opinion. 

\begin{etaremune}
    \item Excellent: Fully objective and faithful to the original; no personal opinions or misrepresentations. 
    \item Good: Mostly objective with minor deviations or interpretive language. 
    \item Satisfactory: Some bias or interpretation present; occasional misrepresentation. 
    \item Poor: Significant bias or misrepresentation; often deviates from the original intent. 
    \item Very Poor: Highly subjective or distorted retelling; misrepresents the original text.
\end{etaremune}

\noindent
3). Grammatical organization and clarity (scale of 1--5) \\
Is the summary well-written, grammatically correct, and logically structured? 

\begin{etaremune}
    \item Excellent: Clear, concise, and free of grammatical errors; logically structured. 
    \item Good: Mostly clear and well-structured with minor grammatical issues. 
    \item Satisfactory: Some grammar or clarity issues that affect readability. 
    \item Poor: Frequent grammatical errors and disorganized structure. 
    \item Very Poor: Incoherent or nearly impossible to understand due to poor grammar or organization.
\end{etaremune}

\noindent
4). Brevity and focus (scale of 1--5) \\
Avoids unnecessary details, succinct and to the point. 

\begin{etaremune}
    \item Excellent: Very concise and focused; no irrelevant information. 
    \item Good: Mostly succinct with minor digressions or redundancies. 
    \item Satisfactory: Some unnecessary detail; could be more focused. 
    \item Poor: Lacks focus; includes much irrelevant or redundant information. 
    \item Very Poor: Rambling or excessively wordy; lacks clear focus.
\end{etaremune}

\textit{Ethics statement}. This study was reviewed by the University of Southern California (USC) Institutional Review Board (IRB) and determined to be exempt from full review §46.104(d)(3). No sensitive information was collected, and all data was de-identified prior to analysis. Participation was voluntary, and informed consent was obtained from all participants prior to data collection.

\subsection{LIDS visualization for LLM summary inference} \label{Subsec.SOFARIvis.supp}

To gain some insights into the intelligence of the LLM summary (e.g., ChatGPT) shown in Section \ref{new.Sec.gptsumminte}, we apply the LIDS visualization suggested in Section \ref{new.Sec.sofarivisual} for more refined LLM summary inference. 
Figure \ref{fig:Utah_Clouds} depicts the $\widehat{k}_j$ latent SVD layers associated with the LIDS summary embedding vector $d_j(\widehat{k}_j)$ given in Section \ref{new.sec.lids.summ.embed} for a representative LLM summary (e.g., ChatGPT) of the Utah article. In particular, each panel of Figure \ref{fig:Utah_Clouds} gives a word cloud plot of selected important key words underlying a specific latent theme (i.e., an SVD layer) of the summary with desired FDR control at level $q = 0.005$. The choice of a smaller FDR level $q = 0.005$ is due to the relatively small standard errors of the SOFARI estimates for the left singular vector components associated with the use of the BERT embeddings, rendering much smaller p-values for significant key words in the latent SVD layers. Consequently, the choice of $q = 0.005$ strikes a good balance between the FDR control and visualization. We set $q = 0.005$ for all LIDS visualization word cloud plots throughout the paper for consistency.

With the aid of Figure \ref{fig:Utah_Clouds}, we can gain some useful insights into the main themes of the LLM summary on the Utah article. See Section \ref{Subsec.SOFARIvis} for the interpretations of the word cloud plots for the corresponding latent SVD layers. 

Indeed, the LIDS visualization for the layered key word selection with controlled FDR is able to uncover and emphasize the most important themes of the text successively, enabling us to know more about its actual contents in a concise visualized fashion.

\subsection{Comparison of different LLMs} \label{Subsec.horizcomp.supp}

\begin{figure}[t]
    \centering
    \includegraphics[width=0.90\textwidth]{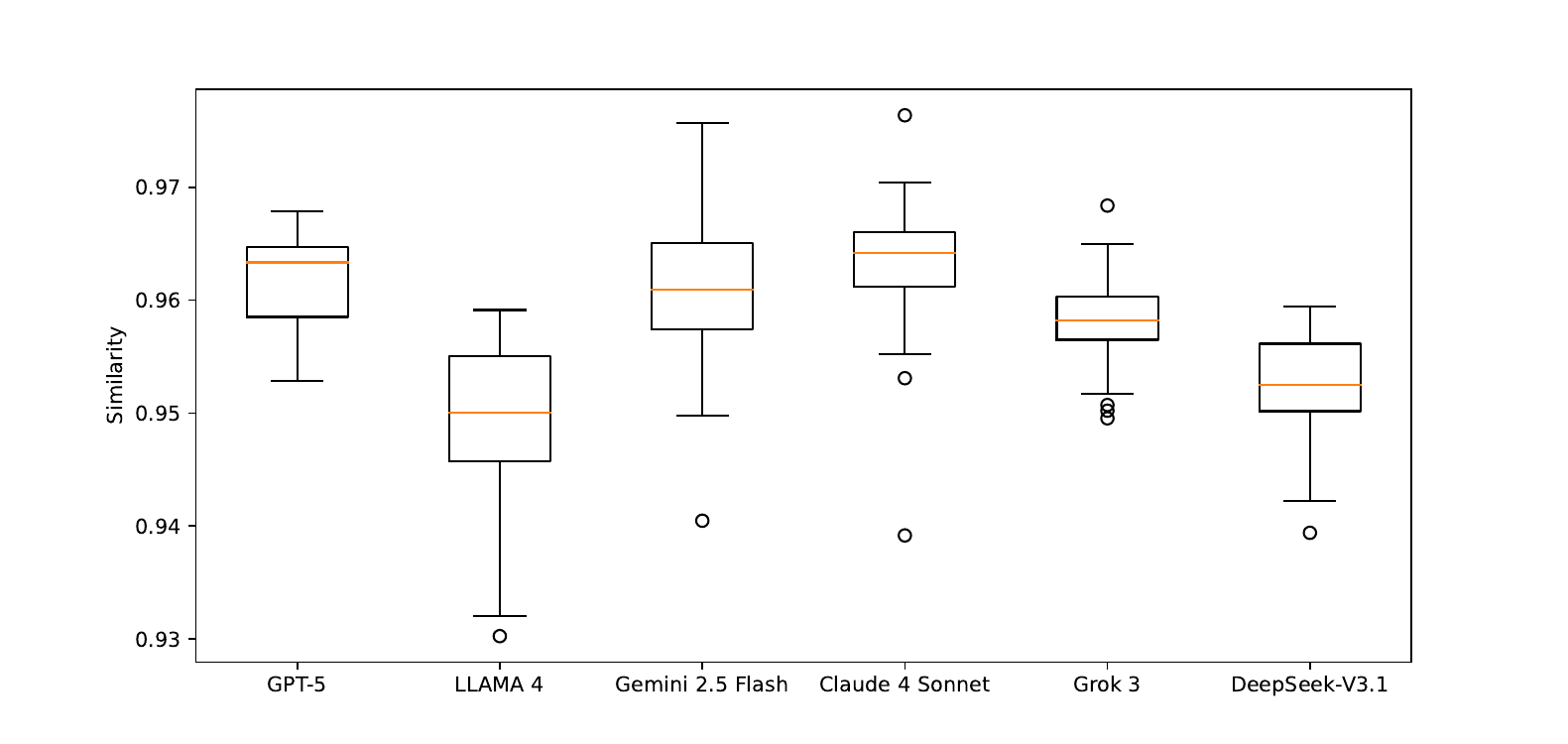}
    \caption{Boxplots of the LIDS similarity measures for different LLMs over $50$ repeated prompts on the Utah article.}
    \label{fig:LLMs_Comp_Box_Utah}
\end{figure}

We present a comparison of different LLMs: ChatGPT, Claude, DeepSeek, Gemini, Grok, and Llama, under the lens of the LIDS similarity measure between the LLM summary and the original text. Figure \ref{fig:LLMs_Comp_Box_Utah} shows all the empirical distributions of the LIDS similarity measures given by different LLMs based on $50$ repeated prompts on the Utah article in terms of the boxplots. By comparing Figures \ref{fig:LLMs_Comp_Box_Utah} and \ref{fig:GPT_Verification_Utah}, it can be seen that all LLM summaries perform well in comparison to the two benchmark summary mechanisms (i.e., the naive summary and random topic summary), since the lowest LIDS similarity measure score for all LLM summaries is above $0.93$, whereas the highest LIDS similarity measure score for the random topic summary is below $0.87$ and that for the naive summary is below $0.70$. Such results indicate that all the LLMs are intelligent in reducing and summarizing large texts. However, there are noticeable differences among the different LLMs in terms of the medians and ranges of the empirical LIDS similarity measures.

\begin{table}[t]
\centering
\caption{Means, standard deviations (SDs), and Sharpe ratios of the LIDS similarity measures for GPT-5 over $50$ repeated prompts and two benchmark summary mechanisms over $50$ random repetitions on the Utah article
}
    \begin{tabular}{lccc} 
        \hline
       Summary Method & Mean & SD & Sharpe Ratio \\
        \hline
        GPT-5 Sum & 0.961594 & 0.003868 & 248.616394 \\
        Naive Sum & 0.675266 & 0.006546 & 103.157628 \\
        Rand Topic Sum & 0.775645 & 0.033374 & 23.240836 \\
        \hline
    \end{tabular} 
    \label{table:GPT_Verification_Utah}
\end{table}

\begin{table}[tp]
\centering
\caption{Means, standard deviations (SDs), and Sharpe ratios of the LIDS similarity measures for different LLMs over $50$ repeated prompts on the Utah article
}
    \begin{tabular}{lccc} 
        \hline
       LLM & Mean & SD & Sharpe Ratio \\
        \hline
        GPT-5 & 0.961594 & 0.003868 & 248.616394 \\
        LLAMA 4 & 0.949213 & 0.006797 & 139.658456 \\
        Gemini 2.5 Flash & 0.960815 & 0.006013 & 159.789622 \\
        Claude 4 Sonnet & 0.963233 & 0.005397 & 178.475634 \\
        Grok 3 & 0.958152 & 0.002849 & 262.873284 \\
        DeepSeek-V3.1 & 0.952481 & 0.004745 & 200.735187 \\
        \hline
    \end{tabular} 
    \label{table:LLM_Comp_Utah}
\end{table}

As mentioned in 
Section \ref{Subsec.horizcomp}, another way to compare different LLMs is to calculate the Sharpe ratio-type measure of the accuracy per unit of uncertainty (i.e., the mean similarity divided by the corresponding standard deviation), exploiting the popularly used Sharpe ratio in finance for ranking stocks and portfolios \citep{Sharpe1966}. As such, we calculate the Sharpe ratios of the LIDS similarity measures for different LLMs as well as the two benchmark summary mechanisms on the Utah article. Tables \ref{table:GPT_Verification_Utah} and \ref{table:LLM_Comp_Utah} summarize the corresponding results in terms of the means, standard deviations, and Sharpe ratios of the LIDS similarity measures. Again, we see from Table \ref{table:GPT_Verification_Utah} that the LLM summary (e.g., ChatGPT) performs much better than the two benchmark summary mechanisms under the Sharpe ratio. In light of Table \ref{table:LLM_Comp_Utah}, it is clear that GPT-5 and Grok 3 are at the top in terms of the Sharpe ratio, indicating their capability and robustness in effectively summarizing the text.

\subsection{Robustness checks across varying text domains}
\label{Subsec.DocDomains.supp}

We finally conduct the robustness checks of LIDS over different text domains: 1) a NASA article, 2) a legal document, and 3) a novel chapter.

\subsubsection{A NASA article} \label{new.sec.nasa}

\begin{figure}[t]
    \centering
    \includegraphics[width=0.95\textwidth]{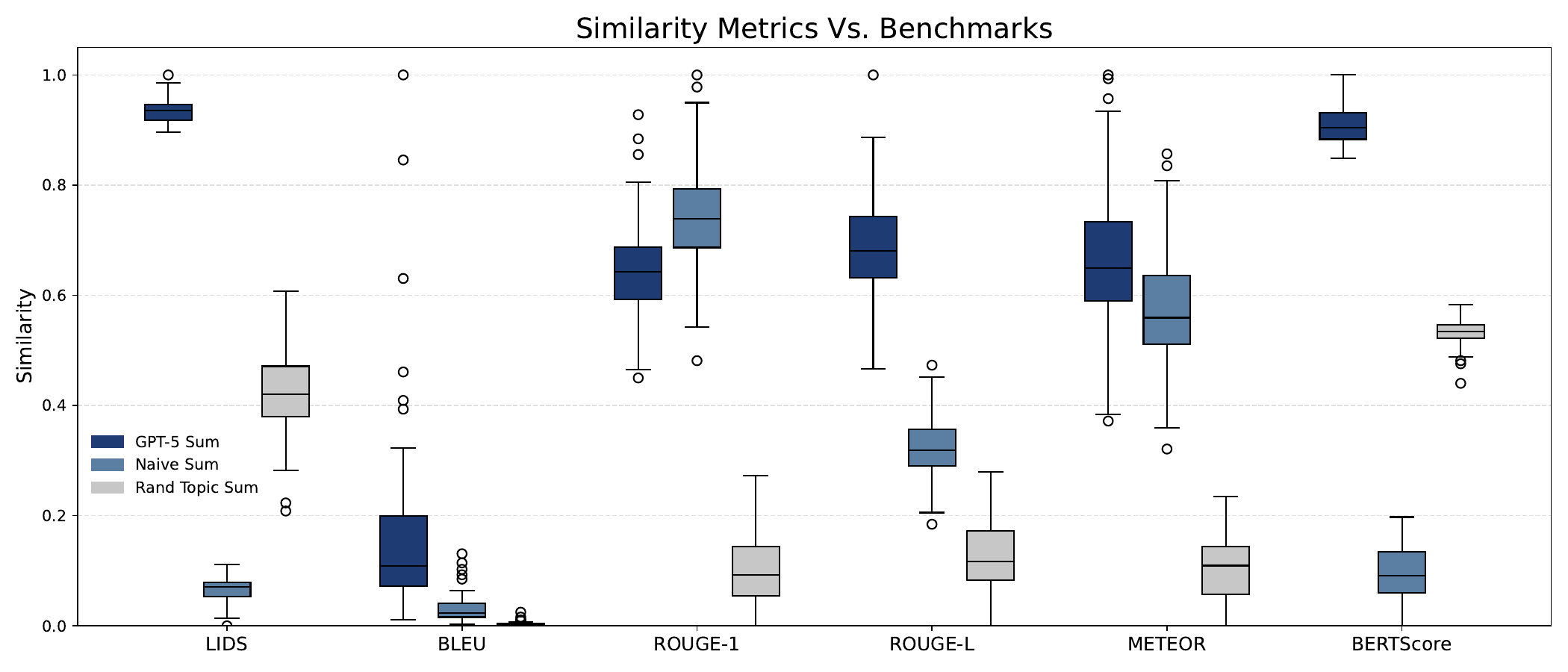}
    \caption{Rescaled boxplots of the LIDS, BLEU, ROUGE-1, ROUGE-L, METEOR, and BERTScore similarity measures for GPT-5 (dark blue) over $50$ repeated prompts and two benchmark summary mechanisms (light blue and gray) over $50$ random repetitions on the NASA article.
    }
    \label{fig:SimMets_Vs_Bench_NASA}
\end{figure}

\begin{figure}[tp]
    \centering
    \includegraphics[width=0.60\textwidth]{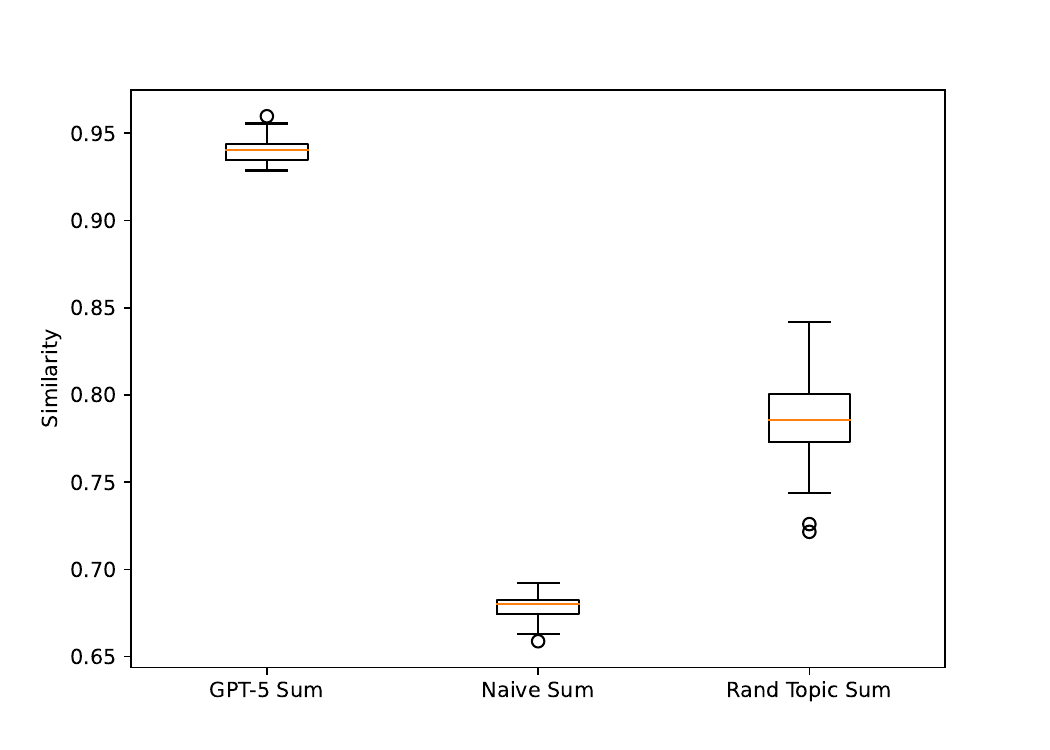}
    \caption{Boxplots of the LIDS similarity measures for GPT-5 over $50$ repeated prompts and two benchmark summary mechanisms over $50$ random repetitions on the NASA article.
    }
    \label{fig:GPT_Verification_NASA}
\end{figure}

\begin{figure}[tp]
    \centering
    \includegraphics[width=0.90\textwidth]{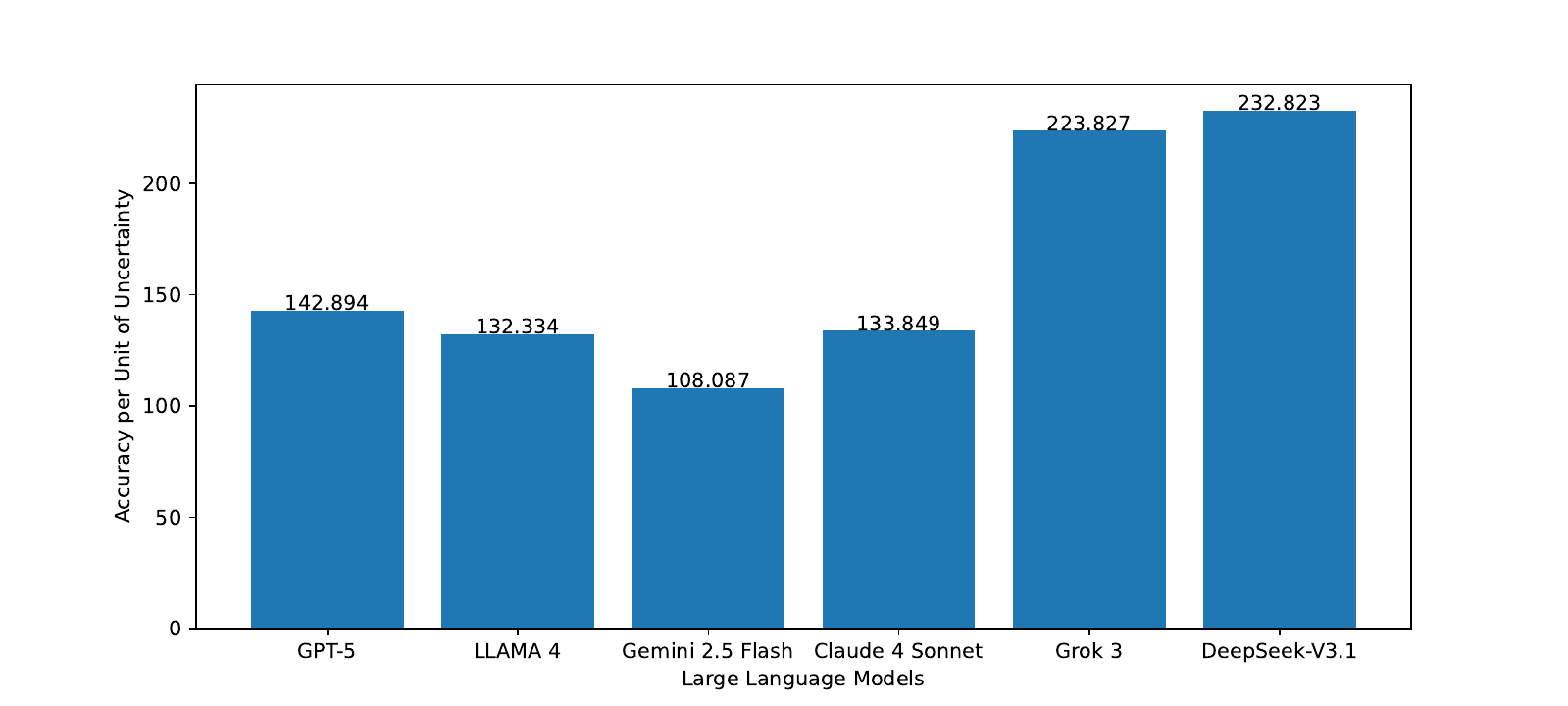}
    \caption{Comparison of different LLMs with LIDS in terms of the Sharpe ratio-type measure of accuracy per unit of uncertainty, i.e., the mean similarity divided by the corresponding standard deviation over $50$ repeated prompts on the NASA article. Larger values indicate better performance.
    }
    \label{fig:LLM_Sharpe_Comp_NASA} 
\end{figure}

\begin{figure}[tp]
    \centering
    \includegraphics[width=0.90\textwidth]{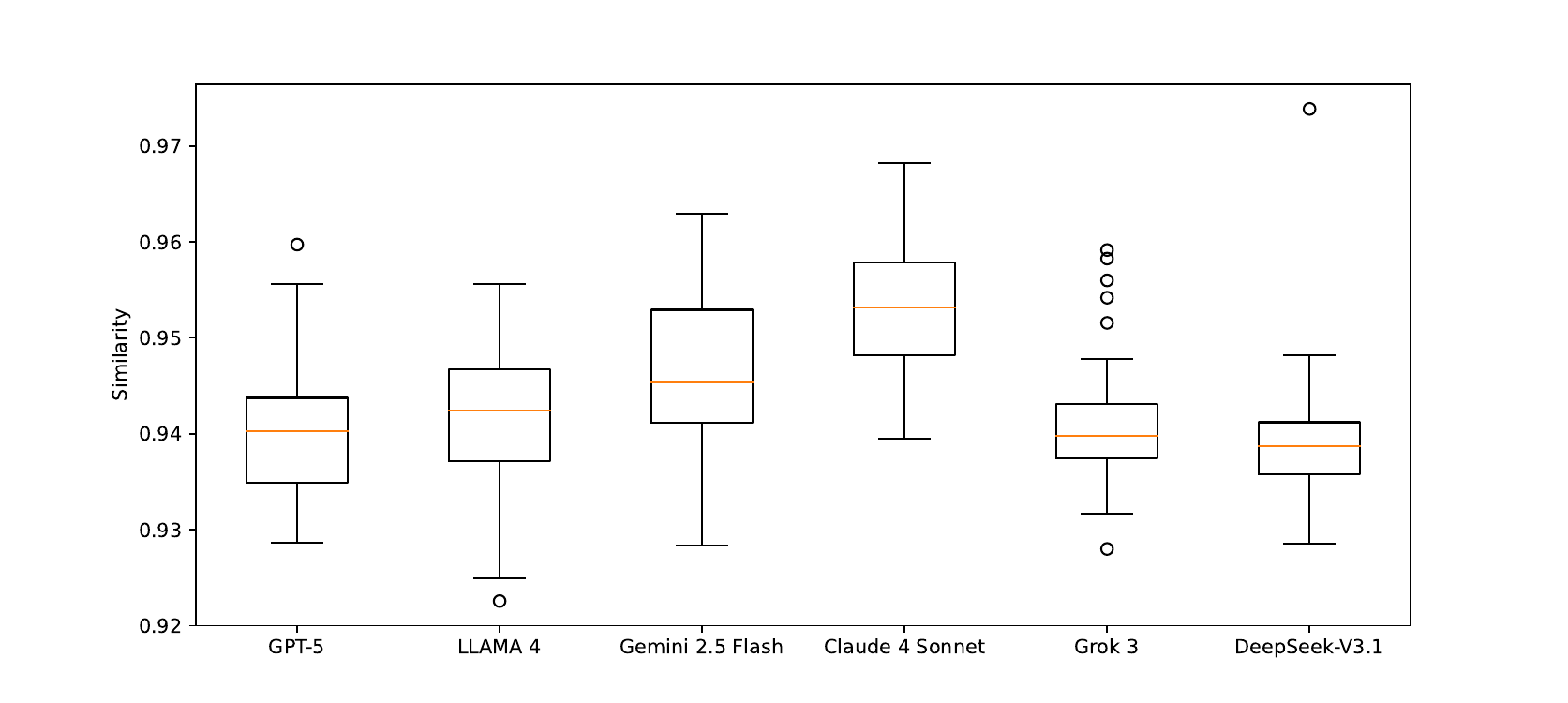}
    \caption{Boxplots of the LIDS similarity measures for different LLMs over $50$ repeated prompts on the NASA article.
    }
    \label{fig:LLMs_Comp_Box_NASA}
\end{figure}

As mentioned in Section \ref{Subsec.DocDomains}, we test the robustness of LIDS on another text application involving the NASA article. 
Figures \ref{fig:SimMets_Vs_Bench_NASA}, \ref{fig:GPT_Verification_NASA}, \ref{fig:LLM_Sharpe_Comp_NASA}, and \ref{fig:LLMs_Comp_Box_NASA} 
are the counterparts of Figures \ref{fig:SimMets_Vs_Bench_Utah}, \ref{fig:GPT_Verification_Utah}, \ref{fig:LLM_Sharpe_Comp_Utah}, and \ref{fig:LLMs_Comp_Box_Utah} 
presented before, with LIDS now applied to the NASA article (as opposed to the Utah article earlier). 
In particular, we see that the performance of LIDS is robust across both the Utah article and the NASA article in effectively evaluating the accuracy and quality of summaries given by different LLMs and the two benchmark summary mechanisms, as well as strong in comparison to several popularly used similarity metrics (e.g., the BLEU, ROUGE-1, ROUGE-L, METEOR, and BERTScore).

\begin{figure}[tp]
    \centering
    \includegraphics[trim={0.85in 0 0 0},clip, width=\textwidth]{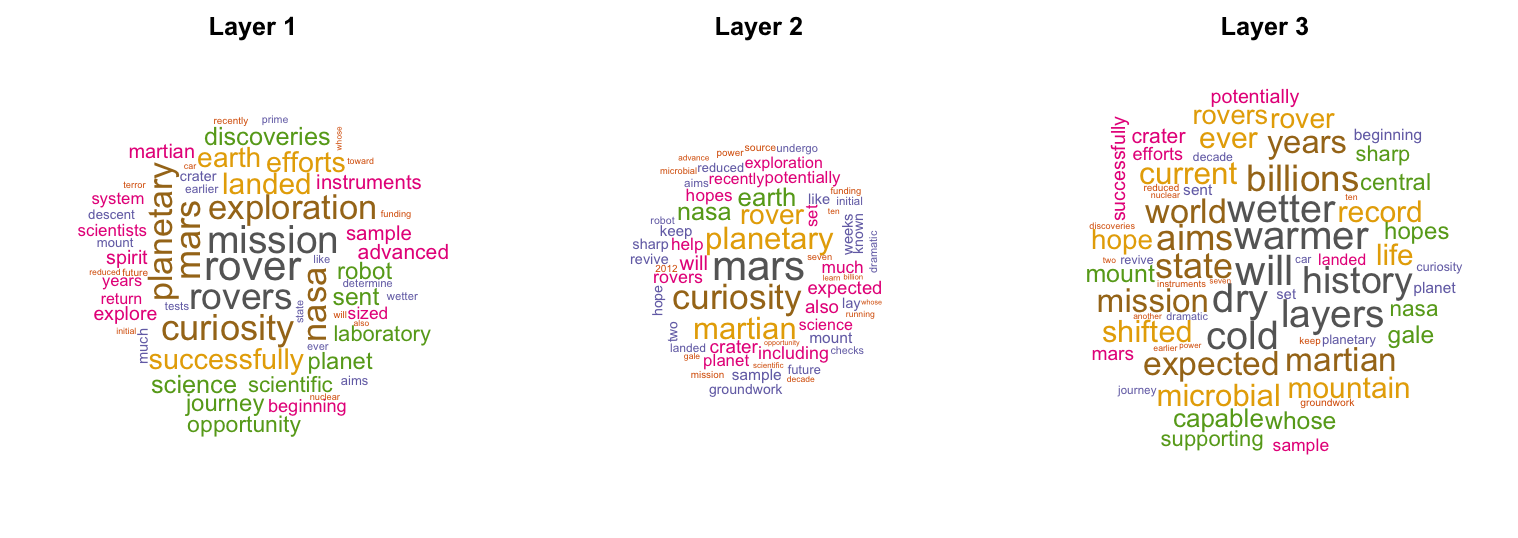}
    \caption{LIDS visualization word cloud plots with FDR control at level $q = 0.005$ for the first three latent SVD layers of a representative LLM summary of the NASA article.}
    \label{fig:NASA_Clouds}
\end{figure}

As in Sections \ref{Subsec.SOFARIvis} and \ref{Subsec.SOFARIvis.supp}, we apply the LIDS visualization suggested in Section \ref{new.Sec.sofarivisual} to gain some insights into the intelligence of the LLM summary (e.g., ChatGPT) on the NASA article. 
Figure \ref{fig:NASA_Clouds} shows the $\widehat{k}_j$ latent SVD layers associated with the LIDS summary embedding vector $d_j(\widehat{k}_j)$ given in Section \ref{new.sec.lids.summ.embed} for a representative LLM summary (e.g., ChatGPT) of the NASA article. In particular, each panel of Figure \ref{fig:NASA_Clouds} provides a word cloud plot of selected important key words underlying a specific latent theme (i.e., an SVD layer) of the summary with controlled FDR at level $q = 0.005$. 
From the layer $1$ word cloud plot in Figure \ref{fig:NASA_Clouds}, we can infer that the NASA article focuses heavily on scientific activity, including experiments and elements of exploration. The accompanying LLM summary references NASA and Mars, suggesting that these experiments and explorations pertain to Martian research. The layer $2$ word cloud plot in Figure \ref{fig:NASA_Clouds} further supports this interpretation, emphasizing themes of Mars exploration, sample collection, and the involvement of a rover. Finally, the layer $3$ word cloud plot in Figure \ref{fig:NASA_Clouds} confirms that the article discusses a Mars rover mission, specifically highlighting the Curiosity rover’s landing on Mars. 
Indeed, the LIDS visualization through layered key word selection with FDR control successively tells us the main theme and subject of the article being about NASA's Curiosity rover on Mars, revealing more about the actual contents of the text under the layered lens.

\begin{figure}[tp]
    \centering
    \includegraphics[width=0.60\textwidth]{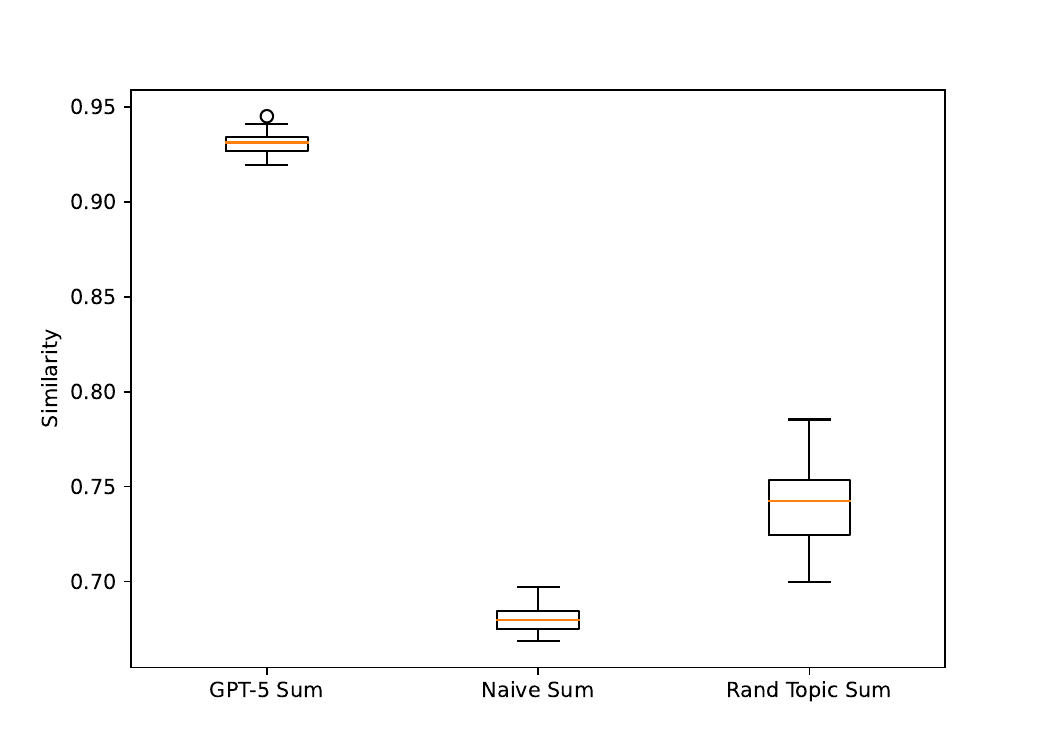}
    \caption{Boxplots of the LIDS similarity measures for GPT-5 over $50$ repeated prompts and two benchmark summary mechanisms over $50$ random repetitions on the legal document (i.e., a school district's alcohol and drug free policy).}
    \label{fig:Legal_Domain}
\end{figure}

\begin{figure}[tp]
    \centering
    \includegraphics[trim={0.85in 0 0 0},clip, width=\textwidth]{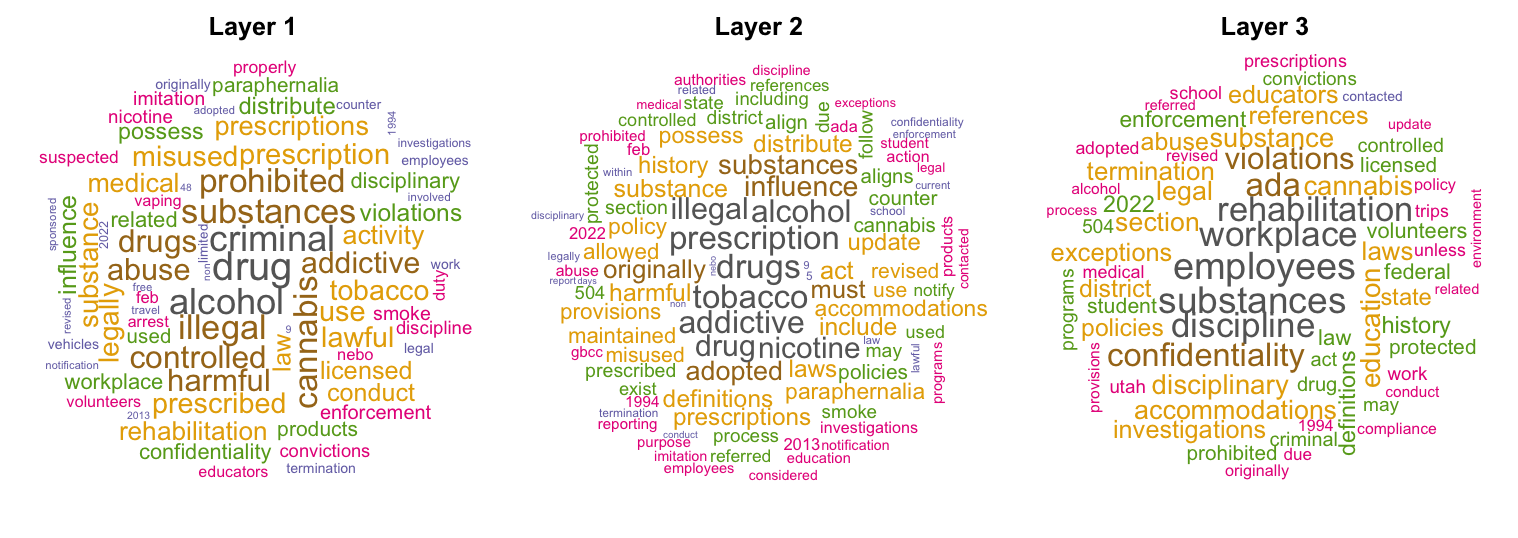}
    \caption{LIDS visualization word cloud plots with FDR control at level $q = 0.005$ for the first three latent SVD layers of a representative LLM summary of the legal document (i.e., a school district's alcohol and drug free policy).}
    \label{fig:Legal_Clouds}
\end{figure}

\begin{figure}[tp]
    \centering
    \includegraphics[width=0.60\textwidth]{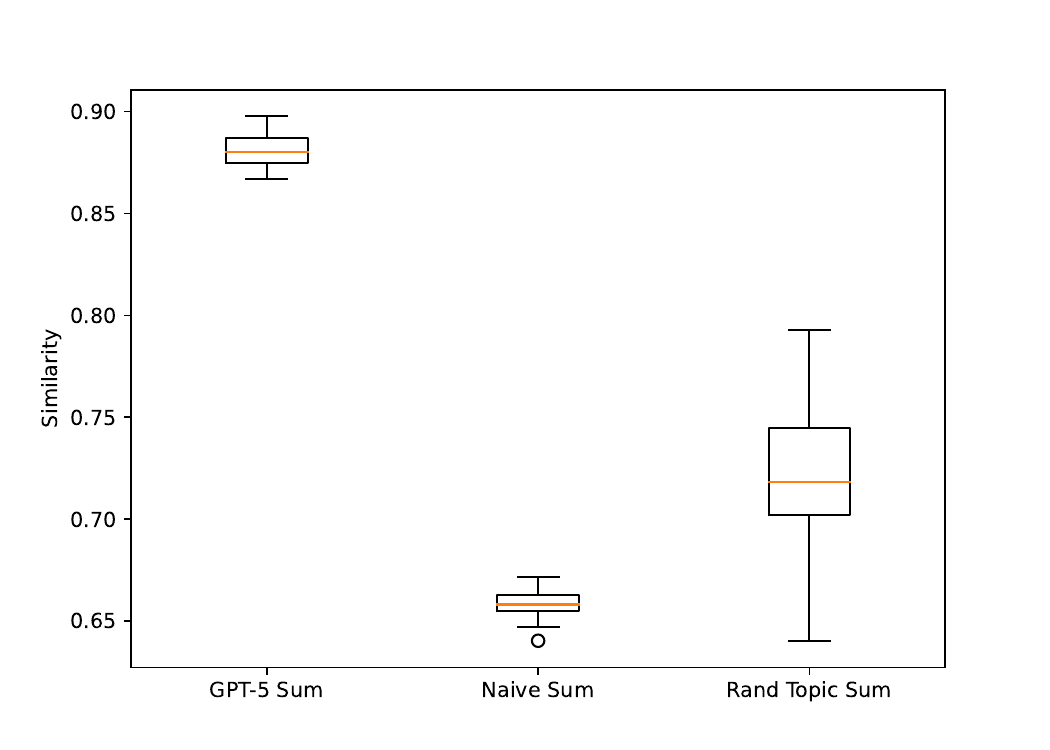}
    \caption{Boxplots of the LIDS similarity measures for GPT-5 over $50$ repeated prompts and two benchmark summary mechanisms over $50$ random repetitions on the novel chapter (i.e., Chapter $3$ of ``Pride and Prejudice'').}
    \label{fig:Story_Domain}
\end{figure}

\subsubsection{A legal document} \label{new.sec.legal}

As mentioned in Section \ref{Subsec.DocDomains}, we further perform the robustness check of LIDS in the legal text domain. Figure \ref{fig:Legal_Domain} presents the results of the LIDS performance on the legal document (i.e., a school district’s alcohol and drug free policy) described in Section \ref{Subsec.DocDomains}. The conclusions are similar in this text domain. We also provide in Figure \ref{fig:Legal_Clouds} the LIDS visualization word cloud plots associated with the three latent SVD layers for a representative LLM summary of the legal document, where the FDR level is set at $q = 0.005$. 
From the layer $1$ word cloud plot in Figure \ref{fig:Legal_Clouds}, we see that the central focus concerns criminal, prohibited, or dangerous substances such as cannabis, tobacco, prescription drugs, and alcohol. There also appears to be discussion of possession and distribution of these substances within a workplace context. In the layer $2$ word cloud plot in Figure \ref{fig:Legal_Clouds}, the emphasis on harmful substances remains, but additional terms related to laws and policies emerge. Notably, smaller words such as student, employee, and education suggest a connection to the educational field. The layer $3$ word cloud plot in Figure \ref{fig:Legal_Clouds} makes this association even clearer. Terms like rehabilitation, discipline, and termination indicate that the text involves workplace policies and legal procedures regarding illegal or harmful substances. Taken together, the evidence strongly suggests that the underlying text pertains to an educational workplace, likely within a school district.

\subsubsection{A novel chapter} \label{new.sec.novel}

We finally conduct the robustness check of LIDS in the book chapter domain as mentioned in Section \ref{Subsec.DocDomains}. Figure \ref{fig:Story_Domain} depicts the results of the LIDS performance on the novel chapter (i.e., Chapter $3$ of ``Pride and Prejudice'') described in Section \ref{Subsec.DocDomains}. 
Furthermore, Figure \ref{fig:PridePrej_Clouds} provides the LIDS visualization word cloud plots associated with the three latent SVD layers for a representative LLM summary of the novel chapter, where the FDR level is set at $q = 0.005$. See Section \ref{Subsec.DocDomains} for the interpretations of these word cloud plots. We continue to see the advantages of LIDS for LLM summary inference in NLP tasks with desired precision and statistical guarantees. 

\end{document}